\def\isarxiv{}

\documentclass[letterpaper]{article} 
\usepackage{aaai2026}  
\usepackage{times}  
\usepackage{helvet}  
\usepackage{courier}  
\usepackage[hyphens]{url}  
\usepackage{graphicx} 
\usepackage{natbib}  
\usepackage{caption} 
\usepackage{algorithm}
\usepackage{algorithmic}

\usepackage{newfloat}
\usepackage{listings}
\DeclareCaptionStyle{ruled}{labelfont=normalfont,labelsep=colon,strut=off} 
\floatstyle{ruled}
\newfloat{listing}{tb}{lst}{}
\floatname{listing}{Listing}
\usepackage{amsmath}
\usepackage{amssymb}
\usepackage{mathtools}
\usepackage{amsthm}
\theoremstyle{plain}
\newtheorem{theorem}{Theorem}[section]
\newtheorem{proposition}[theorem]{Proposition}

\theoremstyle{definition}

\theoremstyle{remark}

\usepackage{booktabs}
\usepackage{fancybox}
\usepackage{color}
\usepackage{graphicx,subcaption}
\usepackage{rotating}
\usepackage{tabularx}
\usepackage{longtable}
\usepackage{cases}
\usepackage{soul}

\newcommand{\RE}{\mathbb{R}}
\newcommand{\RN}[1]{%
  \textup{\expandafter{\romannumeral#1}}%
}

\newcommand{\qedblack}{\hfill\blacksquare}

\title{On the Information Processing of One-Dimensional \\ Wasserstein Distances with Finite Samples}
\author{
    Cheongjae Jang\textsuperscript{\rm 1}, Jonghyun Won\textsuperscript{\rm 1}, Soyeon Jun\textsuperscript{\rm 2,3}, \\
    Chun Kee Chung\textsuperscript{\rm 3}, Keehyoung Joo\textsuperscript{\rm 4}, Yung-Kyun Noh\textsuperscript{\rm 1,4}
}
\affiliations{
    \textsuperscript{\rm 1}Hanyang University, Korea\\
    \textsuperscript{\rm 2}Icahn School of Medicine at Mount Sinai, USA\\
    \textsuperscript{\rm 3}Seoul National University, Korea\\
    \textsuperscript{\rm 4}Korea Institute for Advanced Study, Korea

}

\usepackage{bibentry}

\ifdefined\isarxiv
  \nocopyright
\fi

\begin{document}

\maketitle

\begin{abstract}
Leveraging the Wasserstein distance—a summation of sample-wise transport distances in data space—is advantageous in many applications for measuring support differences between two underlying density functions. 
However, when supports significantly overlap while densities exhibit substantial pointwise differences, it remains unclear whether and how this transport information can accurately identify these differences, particularly their analytic characterization in finite-sample settings.
We address this issue by conducting an analysis of the information processing capabilities of the one-dimensional Wasserstein distance with finite samples. 
By utilizing the Poisson process and isolating the rate factor, we demonstrate the capability of capturing the pointwise density difference with Wasserstein distances and how this information harmonizes with support differences.
The analyzed properties are confirmed using neural spike train decoding and amino acid contact frequency data. The results reveal that the one-dimensional Wasserstein distance highlights meaningful density differences related to both rate and support.
\end{abstract}

\begin{links}
    \link{Code}{https://github.com/cheongjae/one-dim-wasserstein}
\end{links}

\section{Introduction}

\label{sec:intro}

\begin{figure}[t]
    \centering
    \begin{subfigure}[b]{0.45\linewidth}
        \centering
        \includegraphics[width=0.63\textwidth]{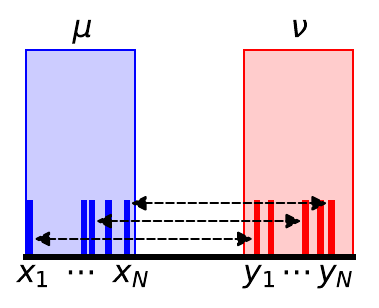}
        \caption{Different support case}
    \end{subfigure}
    \quad
    \begin{subfigure}[b]{0.45\linewidth}
        \centering
        \includegraphics[width=0.63\textwidth]{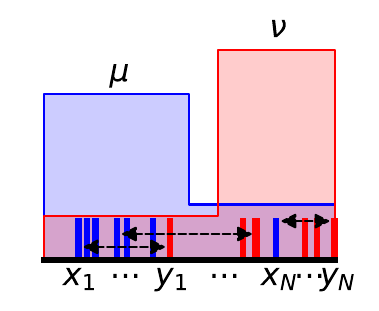}
        \caption{Overlapping support case}
    \end{subfigure}
    \caption{Sample transport between empirical distributions derived from the underlying one-dimensional distributions $\mu$ and $\nu$. 
    Blue and red spikes represent samples drawn from $\mu$ and $\nu$, respectively, with sample transport distances illustrated by dotted arrows.
    In (a), the prominent difference is in the support of $\mu$ and $\nu$, while in (b), the two densities are significantly different despite having the same support. In (b), it is desirable for the measure derived from the sample transport between empirical distributions to represent the pointwise density difference in the underlying distributions.}
    \label{fig:density_diagram}
\end{figure}

The Wasserstein distance, which leverages transport distances between samples, has become a popular tool across various domains for capturing differences between probability density functions \cite{rubner2000earth, haker2004optimal, kusner2015word, schiebinger2019optimal, sihn2019spike}. 
In particular, this metric is known to address support differences due to its reliance on sample transport distances. 
However, it remains unclear whether and how this transport information can appropriately identify pointwise density differences, particularly in terms of their analytic characterization in the finite-sample setting.
This ambiguity persists even in one-dimensional spaces, where the presence of an analytic form for optimal transport simplifies the analysis of Wasserstein distance properties \cite{villani2009optimal, santambrogio2015optimal, peyre2019computational, bobkov2019one} (see Figure~\ref{fig:density_diagram}).

In this paper, we explore the information processing capabilities of the one-dimensional Wasserstein distance between empirical measures derived from sequences of randomly generated finite points, where rates are proportional to the underlying density. 
By employing Poisson processes that depend exclusively on the rate parameter, we derive that the expected transport distance between samples adequately identifies rate differences which can be equivalently understood as measuring the pointwise density difference. 
Furthermore, introducing a shift to the support of one empirical distribution, disregarding any other influencing factors, shows that the one-dimensional Wasserstein distance seamlessly accommodates both positional and rate differences.

Traditionally, the mechanism of encoding rate information and position information have been considered mutually exclusive, particularly in brain research. This dichotomy resembles the long-lasting debate on information processing in the brain: temporal coding versus rate coding. Given the necessity of using density difference information, KL-divergence can appropriately quantify rate differences while is overly sensitive to support differences. Conversely, Wasserstein distance has been suggested as a pragmatically useful alternative that avoids such sensitivities \cite{arjovsky2017wasserstein, ozair2019wasserstein}. However, questions remain about whether Wasserstein distances can reliably treat rate differences, similar to how KL-divergence does, and most importantly, if the distance is minimized when samples come from equivalent densities.

Our examination of the one-dimensional Wasserstein distance reaffirms its practical value for applications requiring the quantification or optimization of differences in rate and support. 
Our discussion extends to the sliced Wasserstein distance---a measure derived as the expectation of the one-dimensional Wasserstein distance applied to projections of multidimensional distributions along arbitrary directions \cite{bonneel2015sliced}. 
These insights could thus be particularly valuable given the recent surge in applications of sliced Wasserstein distances \cite{kolouri2018sliced, deshpande2018generative, lee2019sliced}.

In the experiments, we substantiate our theoretical findings with case studies involving neural spike train decoding and amino acid contact frequency data, showing that sample transport distances and the resulting one-dimensional Wasserstein distance can effectively capture density differences related to both rate and support, leading to enhanced classification performance and data representation. 
These findings highlight the potential utility of the Wasserstein distance-based analysis across diverse fields, particularly in neuroscience and molecular biology.

The paper is organized as follows.
Section~\ref{sec:background} provides background on the one-dimensional Wasserstein distance.
Section~\ref{section:main} presents the derivations of the expected Wasserstein distance using two Poisson processes. 
Section~\ref{sec:experiments} reports the experimental results that verify our findings and show their behavior beyond the Poisson setting.

This article is an extended version of our paper with the same title, accepted to the proceedings of AAAI 2026 \cite{jang2026information}. It includes full proofs, additional examples, and supplemental explanations.

\section{Background}
\label{sec:background}
The $p$-Wasserstein distance between two probability density functions $\mu$ and $\nu$ over $\RE^d$ is defined as
\begin{align}
    W_p(\mu, \nu) = \left( \inf_{\gamma \in \Gamma}\int_{\RE^d} \int_{\RE^d} \|x - y\|^p d\gamma(x,y) \right)^{1/p},
    \label{eqn:p-wass}
\end{align}
where $\Gamma$ denotes the set of all joint distributions on $\RE^d\times \RE^d$ that have respective marginals $\mu$ and $\nu$.

In the one-dimensional case ($d=1$), this admits a closed-form expression using cumulative distribution functions (CDFs) $P$ and $Q$ of $\mu$ and $\nu$, respectively \cite{dall1956sugli, peyre2019computational}.
Specializing to $p=1$ gives
\begin{align}
    W_1(\mu, \nu) = \int_{\RE} |P(u) - Q(u)| du.
    \label{eqn:W1_closed_form}
\end{align}
This distance is the primary focus of our study and is denoted as the Wasserstein distance or $W(\mu, \nu)$ throughout.

Let $\hat{\mu}_N$ and $\hat{\nu}_N$ denote empirical distributions constructed from i.i.d.\ samples $x_1, \ldots, x_N \sim \mu$ and $y_1, \ldots, y_N \sim \nu$, respectively: 
$\hat{\mu}_N = \frac{1}{N} \sum_{i=1}^N \delta_{x_i}$ and $\hat{\nu}_N = \frac{1}{N} \sum_{i=1}^N \delta_{y_i}$, where $\delta_x$ is a Dirac measure at $x \in \RE$. 
Assuming ordered samples ($x_i< x_{i+1}$ and $y_i < y_{i+1}$ for $i=1,\ldots,N-1$), the Wasserstein distance between $\hat{\mu}_N$ and $\hat{\nu}_N$ is the average transport distance between ordered samples (see Appendix~\ref{appendix:one-dim_wass} for details): 
\begin{eqnarray*}
    W(\hat{\mu}_N,\hat{\nu}_N) = \frac{1}{N}\sum^N_{k=1}| x_k - y_k|.
    \label{eqn:W1_empirical}
\end{eqnarray*}

Under mild conditions, $\mathbb{E}[W(\hat{\mu}_N, \mu)] \to 0$ as $N \to \infty$ \cite{boissard2014mean, fournier2015rate, weed2019sharp}, and hence, $\mathbb{E}\left[W(\hat{\mu}_N, \hat{\nu}_N)\right] \to W(\mu, \nu)$ \cite{sommerfeld2019optimal}.

In scenarios involving infinite samples, there exist cases in which the Wasserstein distance can adeptly reveal differences in pointwise densities (or rates). 
For instance, consider two uniform distributions $\mu = \mathcal{U}[0, 1/\lambda_1]$ and $\nu = \mathcal{U}[0, 1/\lambda_2]$ with constant probability densities $\lambda_1$ and $\lambda_2$ for $\lambda_1 \leq \lambda_2$. 
As $N \to \infty$, the Wasserstein distance between the corresponding empirical distributions $\hat{\mu}_N$ and $\hat{\nu}_N$ can be derived analytically according to \eqref{eqn:W1_closed_form} as
\begin{align*}
    &\lim_{N\rightarrow \infty} \mathbb{E}\left[W(\hat{\mu}_N, \hat{\nu}_N)\right] = W(\mu, \nu) 
    = \frac{1}{2} \left(\frac{1}{\lambda_1} - \frac{1}{\lambda_2}\right). 
\end{align*}

However, this analytic solution cannot be straightforwardly extended to the finite sample case. 
The finite sample expectations are instead equal to the sum of the expected sample distances, given as:
\begin{eqnarray}
    \mathbb{E}[W(\hat{\mu}_N,\hat{\nu}_N)] = \frac{1}{N}\sum^N_{k=1}\mathbb{E}[|x_k - y_k|].\label{eq:wass_sample_sum}
\end{eqnarray}
Although the asymptotic behavior implied by \eqref{eqn:W1_closed_form} offers some insight, it remains implicit. The finite-sample expression in \eqref{eq:wass_sample_sum}, {\it per se}, does not explicitly clarify whether and how the expected sample distances $\mathbb{E}[|x_k - y_k|]$ or the empirical Wasserstein distance effectively capture rate differences in finite-sample cases.
This paper aims to characterize these aspects both analytically and empirically.

\section{Rate and Support Difference Encoding in Wasserstein Distance with Finite Samples} \label{section:main}
To directly examine how sample-level transport distances encode pointwise density and support differences in finite-sample settings, we utilize Poisson processes as a tractable analytic framework.
These processes are commonly employed to model the occurrence of random events over time, such as neural spike events  \cite{spike_Poisson_1, spike_Poisson_2}. 
The rate parameter $\lambda > 0$, the defining characteristic of these processes, directly controls the frequency of events (higher $\lambda$ yields denser spikes). 
By focusing on this process, we can effectively isolate the impact of rate and support differences while keeping other variables controlled.

We analyze the rate difference encoding in Section~\ref{sec:rate_encoding} and support difference encoding in Section~\ref{section:delta} for constant rate parameters, followed by a discussion of the time-varying rate parameter case in Section~\ref{sec:time_varying}.

\subsection{Rate Difference Encoding}
\label{sec:rate_encoding}

Consider two sequences of event (or `spike') times, $\{x_i\}_{i=1}^N$ and $\{y_i\}_{i=1}^N$, generated from Poisson processes with constant rates $\lambda_1$ and $\lambda_2$, where $\lambda_1 < \lambda_2$. 
For simplicity and clarity, we begin our analysis with constant rates and identical sample sizes ($N$), though the derivations can be extended to cases with different sample sizes and time-varying rates.

In a Poisson process with a constant rate $\lambda$, the $k$-th event time follows an Erlang distribution  with shape $k$ and rate $\lambda$, having density $p(z; k, \lambda) = \frac{1}{(k-1)!} \lambda^k z^{k-1} e^{-\lambda z}$.
Thus, for our sequences, $x_k \sim p(x_k; k, \lambda_1)$ and $y_k \sim p(y_k; k, \lambda_2)$.

\begin{figure}[t]
\centering
\includegraphics[width=.67\columnwidth]
{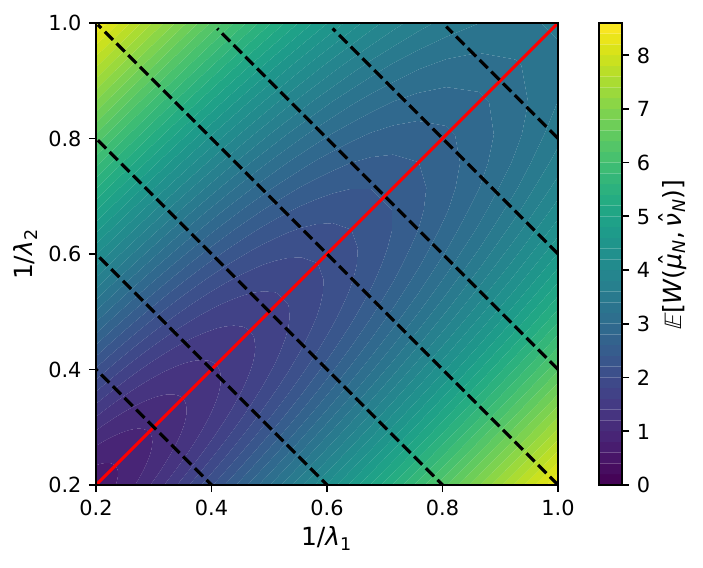}
    \caption{$\mathbb{E}[W(\hat{\mu}_N, \hat{\nu}_N)]$ for $\lambda_1, \lambda_2 \in [1, 5]$ and $N=20$. When the harmonic mean of the rates is constant (black dashed lines), $\mathbb{E}[W(\hat{\mu}_N, \hat{\nu}_N)]$ shows its minimum where the rates are equal ($\lambda_1 = \lambda_2$, red solid line).}
\label{fig:expected_wass}
\end{figure}

Our goal is to analyze the expected Wasserstein distance between the empirical distributions of $\{x_i\}_{i=1}^N$ and $\{y_i\}_{i=1}^N$ as defined in \eqref{eq:wass_sample_sum}.
To do this, we first determine the expected distance between the $k$-th spikes $x_k$ and $y_k$:
{\small
\begin{align*}
    \mathbb{E} [|x_k - y_k|]  = \int_0^\infty \int_0^\infty |x_k - y_k| p(x_k;k,\lambda_1) p(y_k;k,\lambda_2) dx_k d y_k.
\end{align*}
}
The following proposition, derived via a non-trivial computation of this expectation, elucidates how the expected distance between event time pairs $(x_k, y_k)$ reflects the difference in their rates, $\lambda_1$ and $\lambda_2$:
\begin{proposition}
\label{prop:expected_distance}
For the $k$-th spikes $x_k$ and $y_k$ obtained from two Poisson processes of rates $\lambda_1$ and $\lambda_2$, respectively, the expection of the distance between $x_k$ and $y_k$ is  
\begin{align}
\mathbb{E} [|x_k - y_k|] 
& = \frac{\lambda_1 + \lambda_2}{2\lambda_1\lambda_2}\mathbb{E}_{i \sim P(i|2k,p)}\left[|i - (2k - i)|\right],
\label{eqn:prop1}
\end{align}
where $p = \lambda_1/(\lambda_1+\lambda_2)$ and $P(i|2k, p) = \binom{2k}{i} p^{i}(1-p)^{2k-i}$ is the binomial distribution with the parameters $2k$ and $p$.
The minimum of $\mathbb{E}[|x_k - y_k|]$ is achieved when $\lambda_1 = \lambda_2$, under the constraint of constant harmonic mean between rates, i.e., $\frac{2\lambda_1\lambda_2}{\lambda_1 + \lambda_2} = C$ for a $C>0$.\\
\textbf{Proof.} The proof is provided in Appendix \ref{proof:expected_distance}.
$\hfill \square$
\end{proposition}
Note that \eqref{eqn:prop1} is expressed solely in terms of the rates $\lambda_1$ and $\lambda_2$, exhibiting explicit symmetry between them. 
Further analysis of the expected distance in Proposition~\ref{prop:expected_distance} reveals that the expected Wasserstein distance in \eqref{eq:wass_sample_sum} is minimized with $\lambda_1 = \lambda_2$, given that $\frac{\lambda_1+\lambda_2}{2\lambda_1\lambda_2}$ is fixed so that $\frac{1}{2}(\mathbb{E}[x_k]+\mathbb{E}[y_k])$ is constant for each $k$ (see Figure~\ref{fig:expected_wass}).
This finding underscores that Wasserstein distances computed from finite samples reliably and effectively capture rate differences.
A numerical validation is provided in Appendix~\ref{appendix:numerical}.

\paragraph{An infinite-sample case discussion.}
The above discussion becomes increasingly apparent with larger $N$. 
As $k \rightarrow \infty$, the expected distance between the $k$-th spikes, $x_k$ and  $y_k$, normalized by $k$, is given by the following proposition:
\begin{proposition}
\label{theorem:limiting}
Let $x_k$ and $y_k$ be the $k$-th spikes from two Poisson processes with rates $\lambda_1$ and $\lambda_2$, respectively, assuming without loss of generality  that $\lambda_1 < \lambda_2$, and define $s_k = |x_k - y_k| / k$. Then, 
\begin{align}
\begin{split}
    \lim_{k\rightarrow \infty} \mathbb{E}\left[s_k\right] & = \frac{1}{\lambda_1} - \frac{1}{\lambda_2} \quad \text{and} 
    \quad \lim_{k\rightarrow \infty} \textup{Var}\left[s_k\right]  = 0.
    \label{eqn:prop2}
\end{split}
\end{align} 
\textbf{Proof.} The proof is provided in Appendix \ref{proof:limiting}.
$\hfill \square$
\end{proposition}

Based on Proposition~\ref{theorem:limiting}, when $\lambda_1 < \lambda_2$, $\mathbb{E}[W(\hat{\mu}_N, \hat{\nu}_N)]$ for large $N$ can be approximated, up to the leading order, as 
{\scriptsize
\begin{align}
    \mathbb{E}[W(\hat{\mu}_N, \hat{\nu}_N)] &\approx \frac{1}{N} \sum_{k=1}^N\left(\frac{k}{\lambda_1} - \frac{k}{\lambda_2}\right) = \frac{N+1}{2} \left(\frac{1}{\lambda_1} - \frac{1}{\lambda_2}\right).
    \label{eqn:wass_inf_sample}
\end{align}
}
The dependence of \eqref{eqn:wass_inf_sample} on the inverse rate difference demonstrates how the Wasserstein distance directly captures rate differences.

Additionally, the expression in \eqref{eqn:wass_inf_sample} can be interpreted through its connection to uniform distributions. 
In a Poisson process with $N$ events occurring by a fixed time $t$, the joint distribution of the event times, $p(x_1, \ldots, x_N)$, is uniform over $0 < x_1 < \ldots < x_N < t$ \cite{gallager2013stochastic}. 
As $N \rightarrow \infty$, the leading-order expression in \eqref{eqn:wass_inf_sample} corresponds to the Wasserstein distance between two uniform distributions $\mu=\mathcal{U}\left[0, (N+1)/\lambda_1\right]$ and $\nu = \mathcal{U}\left[0, (N+1)/\lambda_2\right]$.

\subsection{Support Difference Encoding} \label{section:delta}
We now explore how differences in support influence the expected Wasserstein distance. 
This allows us to understand how rate and support difference information are harmonized, which is crucial for a comprehensive understanding of the properties of the Wasserstein distance.
This also aligns with recent findings emphasizing the importance of time encoding over rate encoding in neural information processing, where absolute distances between spikes play a critical role \cite{vanrullen2005spike, gutig2014spike}.

To analyze the effects of support differences, we consider the shift of support by $\Delta t \geq 0$, to the spike timings in one of the Poisson processes. 
We calculate the expected distance between $x_k + \Delta t$ and $y_k$ using a similar method described in Section~\ref{sec:rate_encoding}. 
A comprehensive derivation is in Appendix~\ref{appendix:temporal_shift}.

For clarity and foundational insight, here we focus on the expected distance between the first spikes of each sequence:
{
\small
\begin{align}
    \mathbb{E}[|x_1 + \Delta t - y_1|] & = e^{-\lambda_2 \Delta t}\left(\frac{\lambda_1+\lambda_2}{\lambda_1\lambda_2} \mathbb{E}_{i \sim P(i|2,p)}[|i-1|] \right) \nonumber \\
&  + \Delta t + (1-e^{-\lambda_2 \Delta t})\left(\frac{1}{\lambda_1} - \frac{1}{\lambda_2} \right).     
\label{eq:deriv_absolute_difference}
\end{align}
}

Compared to the rate difference encoding in \eqref{eqn:prop1} for $k=1$,  \eqref{eq:deriv_absolute_difference} introduces additional components: $\mathbb{E}[|x_1 - y_1|]$ weighted by $e^{-\lambda_2 \Delta t}$, the support shift $\Delta t$, and a term involving the inverse rate difference.
When $\Delta t = 0$, the formula recovers the pure rate-based expectation in~\eqref{eqn:prop1} with $k=1$.
As $\Delta t \to \infty$, the expression simplifies to $\Delta t + \frac{1}{\lambda_1} - \frac{1}{\lambda_2}$, where the shift dominates.
A numerical validation of the above result is provided in Appendix~\ref{appendix:numerical}.

This analysis reveals how the Wasserstein distance can encode both rate and support differences and how these factors jointly influence the distance. 
The balance between these two types of information---rate difference and support difference---is moderated by the magnitude of the shift $\Delta t$.

\subsection{Discussion on Time-Varying Rate Cases}
\label{sec:time_varying}
Analyzing the case of time-varying rates provides valuable insight into how sample distances can reflect rate and support differences for arbitrary probability distributions.
For the case of nonhomogeneous Poisson processes with a time-varying rate parameter $\mu(t)$, the distribution of the $k$-th spike arrival time is obtained as follows \cite{gallager2013stochastic}: 
\begin{align}
p(x_k;\mu(\cdot)) = \frac{\mu(x_k)}{(k-1)!}\exp(-m(x_k)) m(x_k)^{k-1},
\end{align}
where $m(x) = \int_{0}^{x} \mu(t) dt$. 

To compute the expectation of $|x_k - y_l|$, where $y_l$ is the $l$-th spike arrival time for nonhomogeneous Poisson processes with a time-varying rate parameter $\nu(t)$, we employ a substitution integral: $x_k \mapsto u = m(x_k)$ and $y_l \mapsto v = n(y_l) = \int_{0}^{y_l} \nu(t) dt$. This leads to the following expression:
{
\small
\begin{align}
\begin{split}
&\int_{0}^\infty \int_{0}^\infty |x_k - y_l| p(x_k; \mu(\cdot)) p(y_l; \nu(\cdot)) dx_k dy_l \\
&= \int_{0}^\infty \int_{0}^\infty |m^{-1}(u) - n^{-1}(v)| \frac{1}{(k-1)!}\exp(-u) u^{k-1} \\
&\cdot \frac{1}{(l-1)!}\exp(-v) v^{l-1}  du  dv.
\end{split}
\label{eqn:time_varying}
\end{align}
}
This formulation allows our derivations to extend to more general cases where the double Laplace transform of $|m^{-1}(u) - n^{-1}(v)| u^{k-1} v^{l-1}$ is analytically tractable. 
In such scenarios, we expect differences in rates and supports to be explicitly captured in the sample transport distances and the resulting Wasserstein distances. 
We use this idea to construct an illustrative example that integrates sliced Wasserstein distances into our framework (see Appendix~\ref{appendix:sliced_wasserstein}).

\section{Experiments}
\label{sec:experiments}

\begin{figure*}[t]
\centering
\begin{subfigure}{.28\textwidth}
    \centering
    \includegraphics[width=0.92\textwidth]{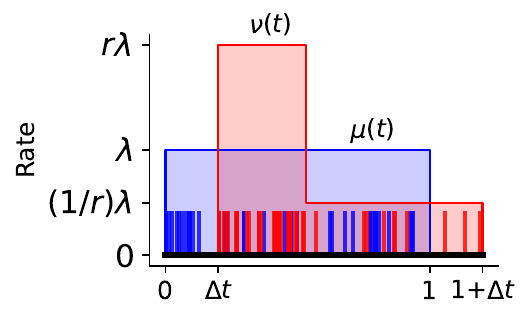}
    \caption{The time-varying rates and samples}
\end{subfigure}\quad
\begin{subfigure}{.63\textwidth}
    \centering
    \includegraphics[width=1.0\textwidth]{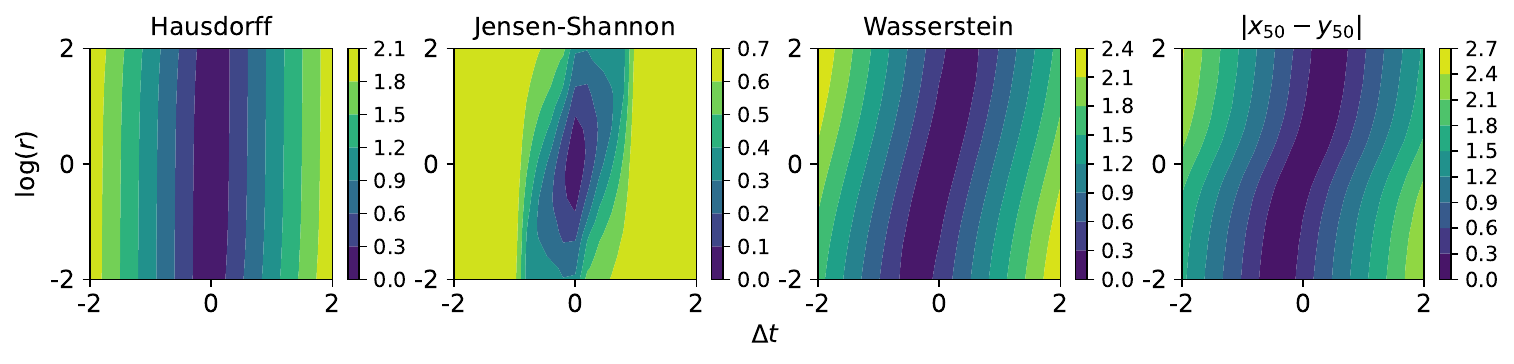}
    \caption{Values between empirical measures with varying $\Delta t$ and $r$}
\end{subfigure}
\caption{A one-dimensional example to compare information processing of the Hausdorff distance, Jensen-Shannon divergence, and Wasserstein distance. 
In (a), blue and red spikes represent the samples generated by Poisson processes with time-varying rates $\mu(t)$ and $\nu(t)$, respectively. In (b), the values are averaged over 1,000 trials.
}
\label{fig:simple_example}
\end{figure*}

To validate our analysis of finite-sample Wasserstein distances, confirm that these analytic insights remain effective beyond the Poisson setting, and assess their practical usefulness in downstream tasks like classification and representation learning, we conduct experiments on synthetic (Section~\ref{sec:synthetic}) and real-world data, including neural spike trains (Sections~\ref{sec:spike_bench_retina} and \ref{sec:spike_train}) and amino acid contacts (Section~\ref{sec:amino_acid}).


\subsection{A Synthetic Data Experiment}
\label{sec:synthetic}

We present a simple example illustrating how sample distances and the Wasserstein distance capture rate and support differences, effectively integrating both. 
For comparison, we also consider alternative measures that capture support or rate differences in distinct ways, such as the Hausdorff distance and the Jensen-Shannon (JS) divergence.

To create a setting with controllable rate and support differences, we use empirical measures from samples generated by two Poisson processes with varying rates $\mu(t)$ and $\nu(t)$, as shown in Figure~\ref{fig:simple_example}(a) and detailed in Appendix~\ref{appendix:synthetic}. 

As depicted in Figure~\ref{fig:simple_example}(b), the Hausdorff distance fails to capture rate differences, while the JS-divergence is overly sensitive with respect to $\Delta t$ when $|\Delta t| < 1$ and then saturates when $|\Delta t| \geq 1$.  
In contrast, the Wasserstein distance and the sample distance, such as $|x_{50} - y_{50}|$, effectively capture and harmonize both rate and support differences.

To quantitatively assess how well sample transport distance captures rate and support differences, we conduct an experiment to predict these differences using transport costs as features. 
For sequences generated from Poisson processes with a constant rate $\mu(t)$ and a piecewise constant rate $\nu(t)$ (defined by two rate ratios, $r_1$ and $r_2$, and a support shift $\Delta t$ relative to $\mu(t)$), we extract ten-dimensional features representing transport costs computed over corresponding decile partitions of the empirical measures.
These features are used to estimate $\log(r_1)$, $\log(r_2)$, and $|\Delta t|$ of $\nu(t)$ by training a three-layer fully connected neural network.

For comparison, we also perform the same task using features derived from (i) the bin-wise divergence values for JS-divergence between the probability mass functions of the samples (computed with ten equal-sized bins), and (ii) two directed Hausdorff distances between the samples.
Further details are in Appendix~\ref{appendix:synthetic}.

\begin{table}[bt]
\caption{$R^2$ scores for estimating rate and support differences in the synthetic example using various features.}
\label{table:synthetic}
\begin{center}
\resizebox{\columnwidth}{!}{%
\begin{sc}
\begin{tabular}{lccc}
\toprule
Feature & $R^2$ for $\log(r_1)$ & $R^2$ for $\log(r_2)$ & $R^2$ for $|\Delta t|$ \\
\midrule
Directed Hausdorff & 43.7$\pm$ 0.5& 43.9$\pm$ 0.3& 70.4$\pm$ 0.3 \\
Bin-Wise JS Divergence & 64.0$\pm$ 0.4& 68.4$\pm$ 0.3& 70.3$\pm$ 0.1\\
Sample Transport Cost & {\bf 81.5} $\pm$ 0.1& {\bf 81.9}$\pm$ 0.2& {\bf 98.9}$\pm$ 0.0 \\
\bottomrule
\end{tabular}
\end{sc}
}
\end{center}
\end{table}

Table~\ref{table:synthetic} presents the $R^2$ values for each feature set.
The results show that sample transport cost features achieve higher $R^2$ scores than other distance-related features, demonstrating their effectiveness in capturing both rate and support differences.
This supports our theoretical analysis that sample transport distance effectively encodes these differences.

\subsection{Retinal Ganglion Cell Stimulus Classification}
\label{sec:spike_bench_retina}
\begin{table}[bt]
\caption{Classification test AUC for retinal stimulus types.}
\label{table:retina}
\begin{center}
\resizebox{\columnwidth}{!}{%
\begin{sc}
\begin{tabular}{lccc}
\toprule
Method & Retina-All & Retina14 & Retina23 \\
\midrule
FCN & 0.945 $\pm$ 7e-04 & 0.962 $\pm$ 7e-04 & 0.925 $\pm$ 3e-04\\
FCN + SD1 & {\bf 0.951} $\pm$ 4e-04 & {\bf 0.971} $\pm$ 9e-04 & {\bf 0.931} $\pm$ 6e-04\\
FCN + SD2 & 0.945 $\pm$ 1e-03 & {\bf 0.968} $\pm$ 1e-03 & {\bf 0.935} $\pm$ 4e-04\\
\midrule
InceptonTime & 0.937 $\pm$ 8e-04 & 0.955 $\pm$ 2e-03 & 0.889 $\pm$ 2e-03\\
InceptonTime + SD1 & {\bf 0.951} $\pm$ 6e-04 & 0.960 $\pm$ 2e-03 & {\bf 0.912} $\pm$ 3e-03\\
InceptonTime + SD2 & {\bf 0.950} $\pm$ 1e-03 & {\bf 0.966} $\pm$ 2e-03 & {\bf 0.913} $\pm$ 1e-03\\
\midrule
ResNet & 0.937 $\pm$ 7e-04 & 0.967 $\pm$ 2e-03 & 0.898 $\pm$ 8e-04\\
ResNet + SD1 & {\bf 0.943} $\pm$ 1e-03 & 0.965 $\pm$ 8e-04 & {\bf 0.912} $\pm$ 1e-03\\
ResNet + SD2 & {\bf 0.948} $\pm$ 5e-04 & 0.967 $\pm$ 2e-03 & {\bf 0.913} $\pm$ 2e-03\\
\midrule
XceptionTime & 0.944 $\pm$ 6e-04 & 0.970 $\pm$ 1e-03 & 0.930 $\pm$ 8e-04\\
XceptionTime + SD1 & 0.947 $\pm$ 2e-03 & {\bf 0.979} $\pm$ 6e-04 & 0.932 $\pm$ 1e-03 \\
XceptionTime + SD2 & {\bf 0.950} $\pm$ 6e-04 & {\bf 0.978} $\pm$ 9e-04 & 0.932 $\pm$ 1e-03 \\
\bottomrule
\end{tabular}
\end{sc}
}
\end{center}
\end{table}
Here, we evaluate the usefulness of information from sample transport distances for stimulus-type classification using retinal ganglion cell spike train data. 
Previous work \cite{lazarevich2023spikebench} showed that using inter-spike interval (ISI) values alone achieves high classification performance for individual spike train chunks. 
We show that complementing ISI values with sample transport distances between their empirical measures—which capture distributional differences—can further improve this performance.
\paragraph{Dataset.} 
We use spike time data from multi-electrode array recordings of salamander retinal ganglion cells under four stimulus types \cite{prentice2016error}. 
The classification tasks include a multiclass problem (Retina-All) to classify all types and binary problems (Retina14, Retina23) to distinguish between specific class pairs.
Following the preprocessing pipeline in \cite{lazarevich2023spikebench}, we construct the dataset $\mathcal{D}=\{(X_i, y_i)\}_{i=1}^T$, where $X_i = (x_{i,1}, \ldots, x_{i,200}) \in \RE^{200}$ represents an ISI temporal vector extracted from single-neuron spike train windows and $y_i \in \{1,2,3,4\}$ is the corresponding label.
The dataset is split into $T_{train}$ training and $T_{test}$ test sets.
\paragraph{Methods.}
To extract features that capture distributional differences in ISI values, we compute the sample transport cost between the empirical measure of each input sequence $X_i=\{x_{i,k}\}_{k=1}^{200}$ and that of the training set $Y=\{x_{i,k}\}_{i=1}^{T_{train}},_{k=1}^{200}$.
By evaluating transport costs over each $1/200$ probability mass interval, we obtain a 200-dimensional feature, called SD1. 
Alternatively, transport costs can be computed between $X_i$ and the class-wise aggregated empirical distributions, yielding another feature set, SD2. 

We train four types of 1D CNN models, such as FCN and ResNet \cite{wang2017time}, InceptionTime \cite{ismail2020inceptiontime}, and XceptionTime \cite{rahimian2019xceptiontime}, using ISI data and SD1 or SD2. 
These inputs are passed through separate CNNs, and the outputs are concatenated for final classification.
SGD is used to minimize the cross-entropy loss. 
See Appendix~\ref{appendix:spikebench} for details.
\paragraph{Results.} 
Table~\ref{table:retina} reports the test AUC for each classification task, with p-values below 0.05 in bold.  
Incorporating SD1 or SD2 features alongside ISI data shows statistically significant improvements in performance over the baseline trained using only ISI data. 
This demonstrates that effectively leveraging distribution differences—e.g., rate and support differences—via sample transport distances can enhance stimulus type classification performance.

\subsection{An Analysis of Human Neural Spike Trains}
\label{sec:spike_train}

\begin{figure*}[t]
\centering
\begin{subfigure}{.31\textwidth}
\centering
\includegraphics[width=1.0\textwidth]{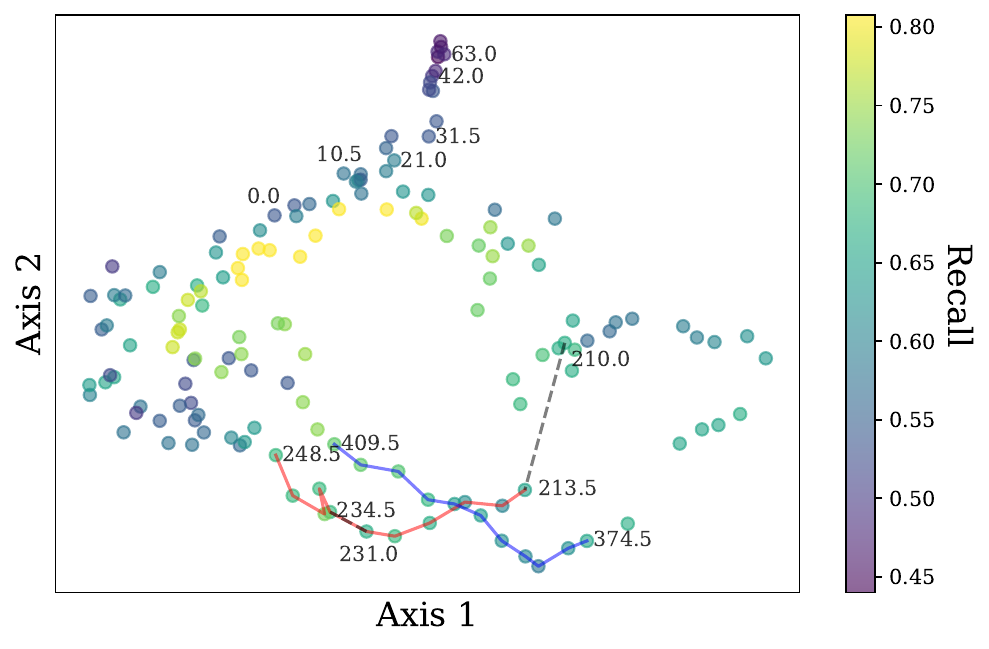}
\caption{Wasserstein distance}
\end{subfigure} \quad
\begin{subfigure}{.31\textwidth}
\centering
\includegraphics[width=1.0\textwidth]{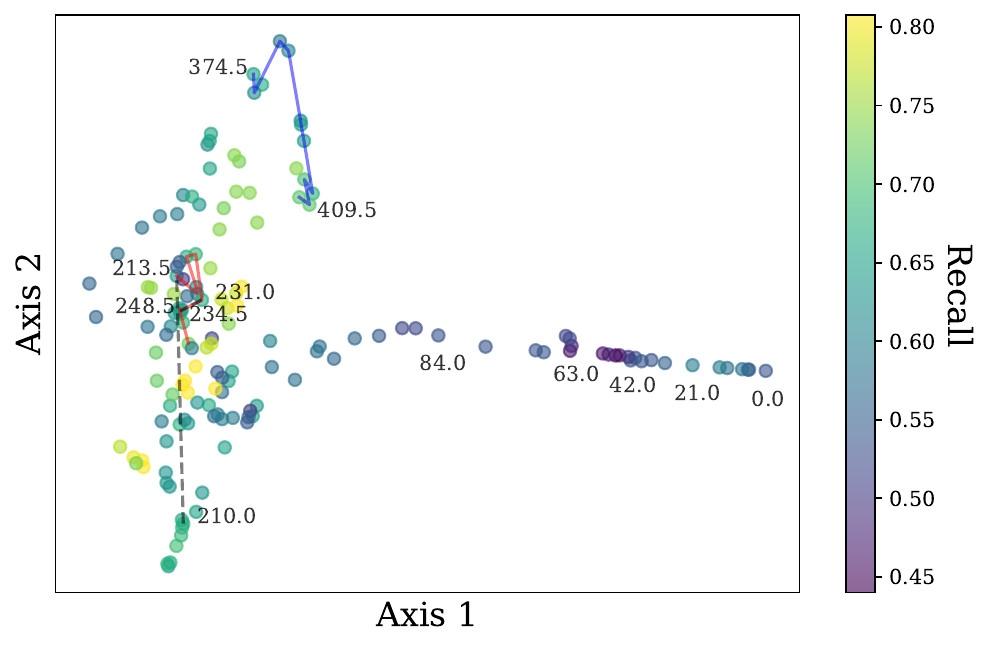}
\caption{Spike count difference}
\end{subfigure} \quad
\begin{subfigure}{.31\textwidth}
\centering
\includegraphics[width=1.0\textwidth]{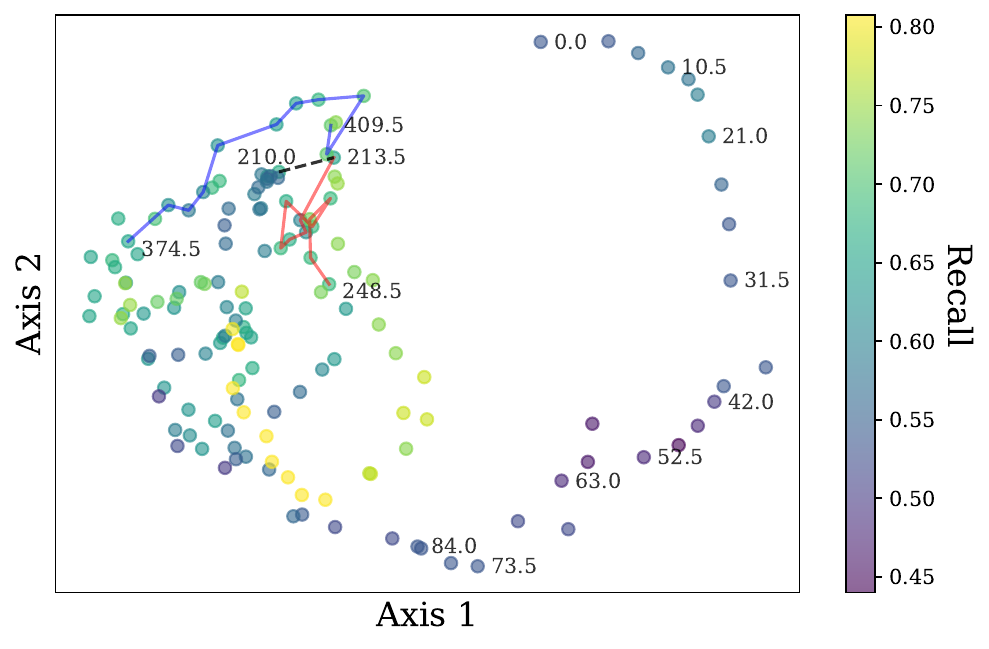}
\caption{Victor-Purpura distance}
\end{subfigure}
\caption{Three Isomap embeddings of human neural spike trains. We present the embedding obtained using the Wasserstein distance in (a), that from the spike count difference in (b), and that from the Victor-Purpura (VP) distance in (c).
}
\label{fig:neural_embedding}
\end{figure*}

\begin{figure*}[t]
\centering
\includegraphics[width=.8\textwidth]
{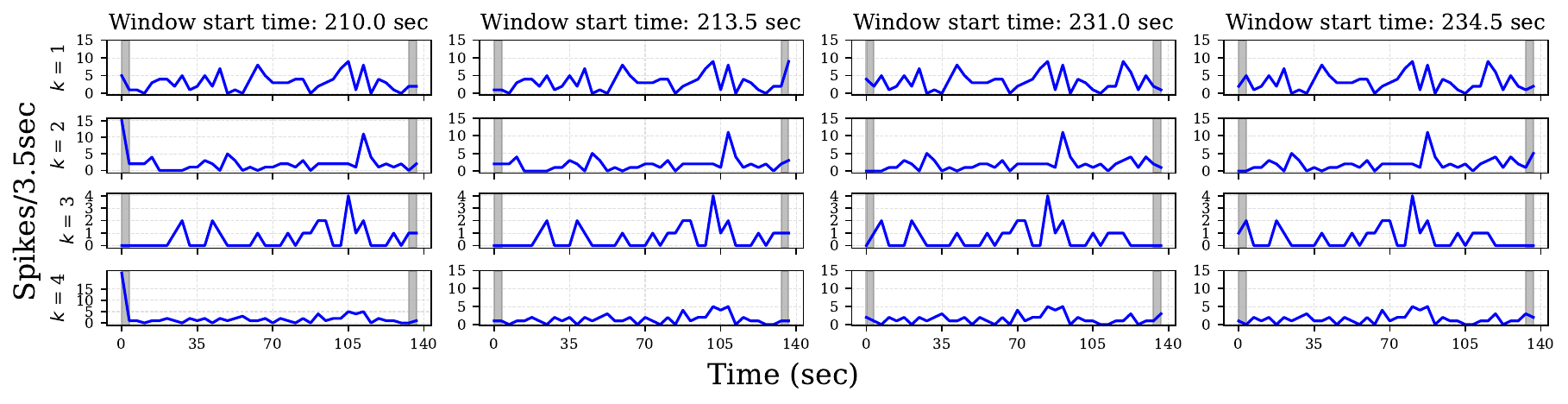}
\caption{Spike count histograms for windows starting at 210.0 seconds, 213.5 seconds, 231.0 seconds, and 234.5 seconds. Shaded areas in each histogram indicate the beginning and end regions of each window. 
}
\label{fig:neural_spike_data}
\end{figure*}

We examine the information processing capabilities of the Wasserstein distance using human neural spike train data. 
By segmenting long spike trains with a sliding window, we explore temporal shifts and rate differences across segments. 
Analyzing embeddings that preserve pairwise distances between spike trains, we show that the Wasserstein distance effectively encodes these variations, offering meaningful insights into the dynamics of individual spike trains. 

\paragraph{Dataset.} 
The neural spike trains used in this study were recorded from the hippocampus of a human subject during the retrieval phase of a word memory task \cite{jun2020task, jun2023hippocampal}.
A 667-sec spike time series was captured simultaneously from four hippocampal microelectrodes and segmented into a series of 140-sec windows with a 3.5-sec sliding interval, yielding a dataset $\mathcal{D} = \{X_{i}\}^{T}_{i=1}$ with $T=151$. 
Each $X_{i} = (X^{(1)}_{i}, X^{(2)}_{i}, X^{(3)}_{i}, X^{(4)}_{i})$ represents a four-channel spike train, where $X^{(k)}_i = \{x_{j}^{(i,k)}\}_{j=1}^{l_{i,k}}$ is the spike train recorded from the $k$-th electrode in window $i$ (see Appendix~\ref{appendix:neural_spike} for details).

\paragraph{Embedding algorithm.} We employ the Isomap algorithm \cite{tenenbaum2000global} to obtain two-dimensional embeddings that aim to preserve pairwise Wasserstein distances between spike trains in $\mathcal{D}$.  
For each four-channel spike train $X_i$, we construct empirical distributions $\hat{\mu}^{(k)}_i$ for each channel $X^{(k)}_i$. 
The composite Wasserstein distance between spike trains is then calculated as $W^*\left(X_{i}, X_{j}\right) = \sqrt{\sum^4_{k=1} W^2(\hat{\mu}^{(k)}_i, \hat{\mu}^{(k)}_j)}$.

For comparison, we also obtain Isomap embeddings using other commonly used dissimilarity measures in neural spike train analysis, including the spike count difference $d(X_i, X_j) = \sqrt{ \sum^4_{k=1} (|X^{(k)}_i| - |X^{(k)}_j|)^2}$ and the Victor-Purpura (VP) distance \cite{victor1997metric}. 
Additional embedding results using a kernel feature space distance from \cite{park2013kernel} are provided in Appendix~\ref{appendix:embedding}.

\paragraph{Rate difference and temporal shift encoding in Wasserstein embeddings.}
Figure~\ref{fig:neural_embedding} shows the Isomap embeddings generated using the Wasserstein distance, spike count difference, and VP distance. 
In each embedding, point color indicates the recall of old words during the memory task for the corresponding time window. 

When examining embeddings obtained from the spike count difference measure and the VP distance, we found that the primary variation only distinguishes the initial windows from the rest, failing to capture meaningful patterns in later windows.
In contrast, the Wasserstein distance embedding goes beyond this limitation. It shows a smoother trajectory (note the red lines representing identical time intervals across subfigures in Figure~\ref{fig:neural_embedding}) and more consistent variations in recall values. 

A closer examination reveals instances of sharp jumps between successive points, the one marked by the black dotted line in Figure~\ref{fig:neural_embedding}(a). 
The two spike count histograms on the left in Figure~\ref{fig:neural_spike_data}, corresponding to this jump, show significant neural activity at the beginning or end of the window. 
This creates a large rate difference between adjacent windows. 
The Wasserstein distance captures this by representing it as a substantial embedding difference, enabling the detection of significant modes in the signal. 
Conversely, with moderate activity and no extreme spikes at either end (as shown in the two histograms on the right of Figure~\ref{fig:neural_spike_data}), temporal shift becomes the primary information, resulting in a continuous trajectory similar to the red and blue lines in Figure~\ref{fig:neural_embedding}(a).
Notably, the VP distance does not effectively capture these distinctions, as Figure~\ref{fig:neural_embedding}(c) shows.

The information processing capabilities of the Wasserstein distance hold strong potential for neuroscience applications. 
By leveraging suitable embedding algorithms, it enables similar neural activity patterns, even those occurring far apart in time, to be represented closely in the embedding space.
An example is the alignment of trajectories such as the red and blue lines in Figure~\ref{fig:neural_embedding}(a), where the subject’s recall performance is comparable at the intersection point.
While these embeddings do not, by themselves, offer definitive conclusions, the analysis suggests that the Wasserstein distance effectively captures both similarities and dissimilarities in neural activity, offering valuable insights for neural signal interpretation.
Importantly, these insights are qualitatively distinct from—and complementary to—those derived from conventional measures such as spike count difference and the Victor–Purpura distance.

\subsection{An Analysis of Amino Acid Contacts}
\label{sec:amino_acid}

\begin{figure*}[t]
\centering
\begin{subfigure}[t]{.39\textwidth}
\centering
\includegraphics[width=0.95\textwidth]{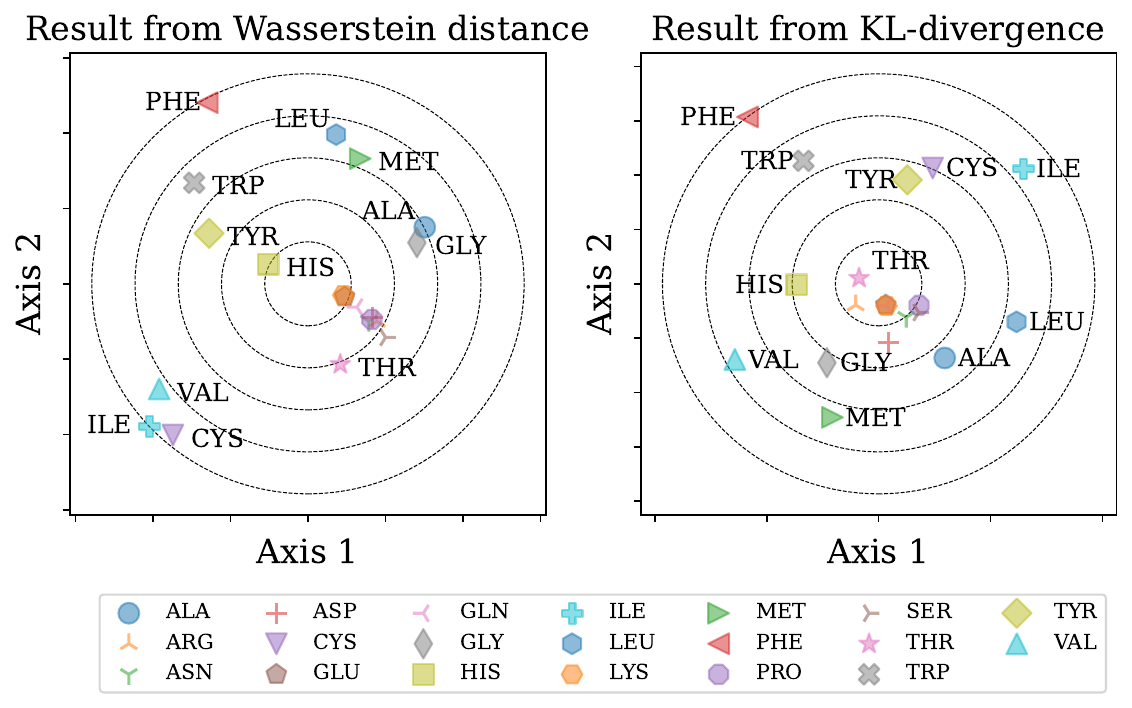}
\caption{Two-dimensional minimum distortion embeddings}
\end{subfigure}\quad
\begin{subfigure}[t]{.21\textwidth}
\centering
\includegraphics[width=0.97\textwidth]{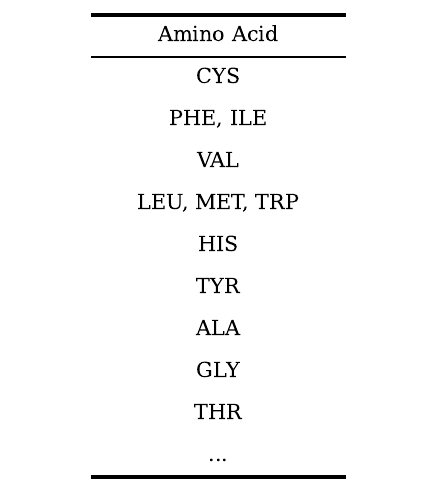}
\caption{A hydrophobicity table}
\end{subfigure}\quad
\begin{subfigure}[t]{.28\textwidth}
\centering
\includegraphics[width=0.95\textwidth]{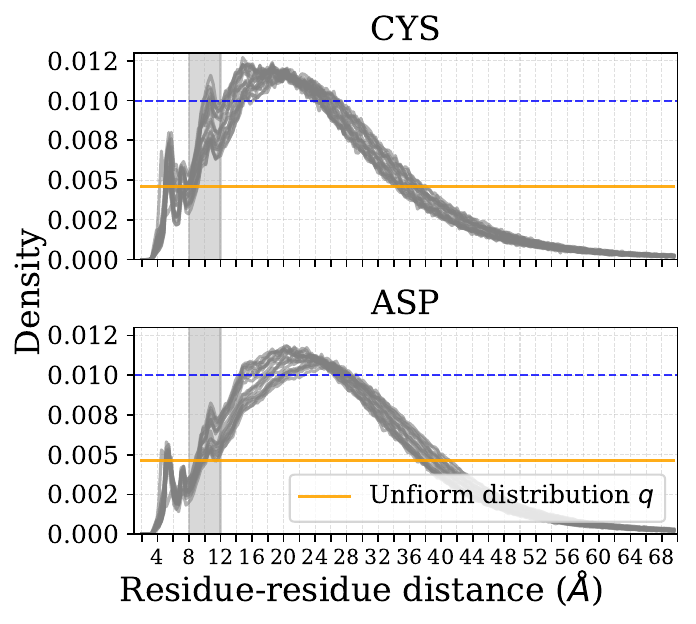}
\caption{$p_{ij}$ of the amino acids CYS and ASP}
\end{subfigure}
\caption{Results from the minimum distortion embeddings of amino acids using pairwise dissimilarities $d_{ij}$. 
In (a), the left shows the two-dimensional embedding obtained from Wasserstein distances, and the right shows that from KL-divergences.
In (b), we present a hydrophobicity rankings of the top twelve hydrophobic amino acids as reported by \cite{rose1985hydrophobicity}.
In (c), shaded areas indicate regions of significant rate differences in $p_{ij}$, with gray curves showing $p_{ij}$ for all $j \neq i$.
}  
\label{fig:protein_map}
\end{figure*}

We analyze amino acid contact data to demonstrate the effectiveness of the Wasserstein distance in capturing rate differences. 
In natural proteins, there are twenty types of amino acids, and the rate at which one amino acid appears near another depends on their spatial distance. 
This rate, known as the contact frequency, is distinct for each amino acid pair.

In this experiment, we use contact frequency data to generate embeddings for the twenty amino acids.
By capturing rate differences via the Wasserstein distance, these embeddings reveal specific properties of amino acids—offering insights that, in some cases, go beyond what is captured by KL divergence in a molecular biology context.

Throughout we use the terms ``amino acid" and ``(protein) residue" interchangeably and denote each amino acid by its standard three-letter code.

\paragraph{Dataset.} Amino acid contact frequencies are represented as a set of pairwise distributions $\{p_{ij} \mid 1 \le i,j \le 20, i\neq j\}$, where each $p_{ij}$ denotes the normalized histogram of spatial distances between the $i$-th and $j$-th amino acids (ordered alphabetically). 
Each histogram is defined over 217 uniformly spaced bins with identical support across all pairs, see Figures~\ref{fig:protein_map}(c) and \ref{fig:protein_distributions} for examples. 
These distributions are derived from 12,508 proteins sourced from the Protein Data Bank (PDB) \cite{berman2000protein}.

\paragraph{Embedding algorithm.} We construct amino acid embeddings based on their pairwise distance distributions $p_{ij}$. 
To define pairwise residue dissimilarity, we use $d_{ij} = \gamma_W \cdot (W(p_{ij}, q) - \beta_W)$, where $q$ is a uniform reference distribution with the same binning as $p_{ij}$, and $\gamma_W, \beta_W > 0$ are scaling factors. 
This dissimilarity reflects the rate difference between $p_{ij}$ and $q$, and it can effectively capture the relative profile of each pairwise distribution $p_{ij}$ compared to others.

We apply $d_{ij}$ to the Riemannian geometric manifold learning algorithm of \cite{jang2021riemannian} to obtain embeddings to preserve geometric structures (e.g., distances and angles) induced from the dissimilarity.
For comparison, we also obtain embeddings using KL-based dissimilarities defined as $d_{ij}= \gamma_{KL}\cdot(D_{KL}(p_{ij}||q) - \beta_{KL})$, with corresponding scaling factors $\gamma_{KL}, \beta_{KL}>0$. 
See Appendix~\ref{appendix:amino_acid} for more details.

\paragraph{Rate differences induced by long-range contacts.} 
Figure~\ref{fig:protein_map}(a) shows two-dimensional embeddings of amino acids based on the Wasserstein distance and KL-divergence. 
In both embeddings, hydrophobic residues such as CYS, ILE, and PHE (as indicated in the hydrophobicity table in Figure~\ref{fig:protein_map}(b)), appear near the periphery, while less hydrophobic residues are positioned closer to the center.
This separation reflects the rate differences between $p_{ij}$ and the reference distribution $q$, suggesting that the contact frequencies of hydrophobic residues deviate more substantially from $q$. 

Further examination of the distributions $p_{ij}$ for CYS in Figure~\ref{fig:protein_map}(c) (and similarly for ILE and PHE in Figure~\ref{fig:protein_distributions} of Appendix~\ref{appendix:amino_acid}) reveals that these larger rate differences, particularly compared to residues in the second row of the figures, are concentrated in the range of 8-12\AA.
From a protein structural perspective, high density in this range indicates a propensity for long-range contacts, a characteristic commonly observed for medium-to-high hydrophobic residues such as CYS, ILE, and PHE \cite{rose1985hydrophobicity, gromiha1999importance, gromiha2004inter}.
Their peripheral positioning in both embeddings (Figure~\ref{fig:protein_map}(a)) effectively highlights these specific properties of amino acids, demonstrating that both Wasserstein and KL-based embeddings capture biologically meaningful features such as long-range contacts.

While both embeddings capture this trend, they differ in how well they align with known hydrophobicity rankings. 
For example, CYS, recognized as the most hydrophobic amino acid by \cite{rose1985hydrophobicity}, appears at the outermost edge in the Wasserstein embedding, but is located closer to the center in the KL-based one.
Overall, the norms of the Wasserstein embeddings show a stronger correlation with the hydrophobicity rankings reported in \cite{rose1985hydrophobicity}:
Kendall’s tau coefficients for radial ordering among the top 10 most hydrophobic residues are 0.722 for Wasserstein versus 0.582 for KL, and for all residues, 0.807 versus 0.731, respectively.
These results highlight the Wasserstein distance’s strength in capturing rate-based structural patterns in amino acid contact data.

\section{Conclusion}
\label{sec:conclusion}

In this paper, we have investigated the information processing capabilities of the one-dimensional Wasserstein distance with finite samples. 
While its ability to capture support differences in density functions is well understood, its mechanism for encoding pointwise density differences through sample transport distances—particularly in finite sample settings—has remained unclear and lacked analytic characterization.
We address this gap by demonstrating rate encoding in one-dimensional distances between empirical measures derived from Poisson processes. 
Our analysis of expected sample distances provides intuitive interpretations and novel insights into how transport distances within samples encode rate differences and integrate them with support variation. 
Furthermore, our analytic and empirical findings highlight the complementary nature of the Wasserstein distance relative to other measures of distributional difference, such as KL-divergence.
We validated these insights through experiments on neural spike train decoding and amino acid contact analysis, demonstrating the broad applicability of Wasserstein-based approaches across diverse domains.

\section*{Acknowledgments}
This work was supported by NRF/MSIT (No. RS-2024-00421203), which supported C. K. Chung and Y.-K. Noh. 
This work was also supported by IITP/MSIT (No. RS-2023-00220628), which supported K. Joo and Y.-K. Noh.
C. Jang was partly supported by NRF/ME (No. RS-2023-00249714).
C. K. Chung was partly supported by NRF/MSIT  (NRF-2021M3E5D2A01019093).
Y.-K. Noh was partly supported by IITP/MSIT (IITP-2021-0-02068, RS-2020-II201373).
We thank the KIAS Center for Advanced Computation for providing computing resources.


\bigskip

\bibliography{references}

\newpage
\appendix
\setcounter{secnumdepth}{2}
\renewcommand\thesubsection{\thesection.\arabic{subsection}}

\onecolumn
\section{Derivations and Proofs}
\label{appendix:derivations}

\subsection{One-Dimensional Wasserstein Distance Between Empirical Measures}
\label{appendix:one-dim_wass}
In Section~\ref{sec:background}, from \eqref{eqn:p-wass}, the one-dimensional 1-Wasserstein distance between $\hat{\mu}_N$ and $\hat{\nu}_N$ is obtained as
\begin{eqnarray}
    W(\hat{\mu}_N,\hat{\nu}_N) := \inf_{\gamma\in\Gamma} \sum^N_{i=1} \sum^N_{j=1} | x_i - y_j| \gamma_{ij},
    \label{eqn:W1_finite_sample}
\end{eqnarray}
where $\Gamma = \left\{\gamma \in \RE_+^{N\times N}| \sum^N_{i=1} \gamma_{ij} = \sum^N_{j=1} \gamma_{ij} = \frac{1}{N}\right\}$. 
Assuming that the samples are ordered, i.e., $x_i< x_{i+1}$ and $y_i < y_{i+1}$ for $i=1,\ldots,N-1$, the optimal transport plan $\gamma$ is given by $\gamma_{ij} = \frac{1}{N} \mathbf{1}_{j = i}$, i.e., $\gamma_{ij}=\frac{1}{N}$ when $i=j$ and zero otherwise \cite{bobkov2019one}.\footnote{When the samples sizes differ ($M \neq N$), the probability masses of $x_i$ and $y_i$ become $1/M$ and $1/N$, respectively. The $\gamma_{ij}$ can be computed in closed form, independent of the sample values, using the Northwest Corner method \cite{hillier2015introduction}.} 
Applying this $\gamma$ to \eqref{eqn:W1_finite_sample} yields the average of the transport distances between ordered samples in \eqref{eqn:W1_empirical}.

\subsection{Proof of Proposition~\ref{prop:expected_distance}}
\label{proof:expected_distance}
\textit{\textbf{Proof.}} Denote by $x_k$ the $k$-th spike from the Poisson process with rate $\lambda_1$ and $y_l$ the $l$-th spike from the Poisson process with rate $\lambda_2$. We derive the expression for the expected distance between $x_k$ and $y_l$ by calculating
\begin{eqnarray}
    \mathbb{E}[| x_k - y_l |] = \int^\infty_0 \int^\infty_0 | x_k - y_l | p(x_k) p(y_l) dx_k dy_l, \label{eq:apen_expec_formal}
\end{eqnarray}
where $x_k, y_l$ respectively follow the Erlang distributions $p(x_k), p(y_l)$ with rates $\lambda_1, \lambda_2$ and shapes $k, l$, i.e.,
\begin{eqnarray}
    p(x_k) &=& \frac{\lambda_1^k x_k^{k-1} \exp(-\lambda_1 x_k)}{(k-1)!}, \label{eqn:erlang_x_k}\\
    p(y_l) &=& \frac{\lambda_2^l y_l^{l-1} \exp(-\lambda_2 y_l)}{(l-1)!}. \label{eqn:erlang_y_l}
\end{eqnarray}
The double integration in \eqref{eq:apen_expec_formal} can be rewritten as 
\begin{eqnarray}
\mathbb{E} [|x_k - y_l|] = 
    \int^\infty_0 \left( \underbrace{\int^\infty_{y_l} (x_k - y_l) p(x_k) dx_k}_{\text{(a)}} + \underbrace{\int^{y_l}_0 -(x_k - y_l) p(x_k) dx_k}_{\text{(b)}} \right) p(y_l) dy_l.  \label{eq:apen_separated}
\end{eqnarray}
When calculating the above integral, the following equation is useful:
\begin{eqnarray}
    \int_a^b x^k\exp(-\lambda x) dx = \left. - \sum^k_{i=0}\frac{k!}{i!\lambda^{k-i+1}}x^i \exp(-\lambda x) \right|_a^b. \label{eq:expon_int}
\end{eqnarray}
Using \eqref{eq:expon_int}, we obtain 
\begin{eqnarray}
    \text{(a)} = \frac{\lambda^k_1}{(k-1)!}\left( \exp(-\lambda_1 y_l) \sum^k_{i=0} \frac{k!}{i!\lambda_1^{k-i+1}}y_l^i - y_l \left( \exp(-\lambda_1 y_l) \sum^{k-1}_{i=0} \frac{(k-1)!}{i!\lambda^{k-i}_1}y_l^i \right)
 \right)
 \label{eqn:integral_a}
\end{eqnarray}
and
\begin{align}
    \text{(b)} = & -\frac{\lambda^k_1}{(k-1)!} \cdot \left(-\exp(-\lambda_1 y_l)\sum^k_{i=0} \frac{k!}{i!\lambda^{k-i+1}_1}y_l^i + \frac{k!}{\lambda_1^{k+1}} \right.  \left. - y_l\left(-\exp(-\lambda_1y_l)\sum^{k-1}_{i=0}\frac{(k-1)!}{i!\lambda^{k-i}_1}y_l^i+ \frac{(k-1)!}{\lambda^k_1} \right) \right).
    \label{eqn:integral_b}
\end{align}
We plug \eqref{eqn:integral_a} and \eqref{eqn:integral_b} into \eqref{eq:apen_separated} and then calculate the integration with respect to $y_l$. After some algebra, we obtain
\begin{eqnarray}
    \mathbb{E} [|x_k - y_l|] = \sum^{k-1}_{i=0} \binom{l-1+i}{i} \frac{2 (k-i) \lambda^{i-1}_1 \lambda^l_2}{(\lambda_1+\lambda_2)^{l+i}} -\frac{k}{\lambda_1} + \frac{l}{\lambda_2}, \label{eq:apen_intermediate}
\end{eqnarray}
which can be rewritten as
\begin{align}
\begin{split}
& \mathbb{E} [|x_k - y_l|] = 
\frac{1}{\lambda_1\lambda_2 (\lambda_1 + \lambda_2)^{k+l-1}} \\
 & \cdot \left[\underbrace{\left(\sum^{k-1}_{i=0}\binom{l-1+i}{i} 2(k-i) \lambda_1^i \lambda_2^{l+1} (\lambda_1 + \lambda_2)^{k-1-i}\right) - \Bigg( (k\lambda_2 - l\lambda_1) (\lambda_1 + \lambda_2)^{k+l-1}\Bigg)}_{\text{(A)}} \right]
 \label{eqn:expected_distance_1}
 \end{split}\\
 & = \frac{1}{\lambda_1\lambda_2 (\lambda_1 + \lambda_2)^{k+l-1}} \sum_{i=0}^{k+l} B_i \lambda_1^i \lambda_2^{k+l-i},
 \label{eqn:expected_distance_2}
\end{align}
where $B_i$ is the coefficient for the $\lambda_1^i \lambda_2^{k+l-i}$ term of (A) in \eqref{eqn:expected_distance_1}.
Using binomial expansion, the coefficient $B_i$ is obtained as 
\begin{numcases}{B_i = }
\sum_{j=0}^i 2(k-i) \binom{l-1+j}{j}\binom{k-j}{i-j}  - k\binom{k+l-1}{i} +l\binom{k+l-1}{i-1} & for $i \leq k-1$ \label{eqn:Bi_firstcase}\\
- k\binom{k+l-1}{i} +l\binom{k+l-1}{i-1} & for $i \geq k$. \label{eqn:Bi_secondcase}
\end{numcases}
Using the identity $\sum_{j=0}^i \binom{l-1+j}{j}\binom{k-j}{i-j} = \binom{k+l}{i}$, which can be proved by an inductive argument, and $ k\binom{k+l-1}{i} - l\binom{k+l-1}{i-1} = (k-i)\binom{k+l}{i}$, we can simplify \eqref{eqn:Bi_firstcase} and \eqref{eqn:Bi_secondcase} as follows:
\begin{equation}
B_i = |k-i|\binom{k+l}{i}.
\label{eqn:Bi_allcase}
\end{equation}

By plugging \eqref{eqn:Bi_allcase} into \eqref{eqn:expected_distance_2}, we obtain
\begin{align}
\mathbb{E} [|x_k - y_l|] 
& = \frac{1}{\lambda_1\lambda_2 (\lambda_1 + \lambda_2)^{k+l-1}} \sum^{k+l}_{i=0} \binom{k+l}{i} \lambda_1^i \lambda_2^{k+l-i} |k-i| \\ 
& = \frac{\lambda_1 + \lambda_2}{\lambda_1\lambda_2}\sum_{i=0}^{k+l} \binom{k+l}{i} \left(\frac{\lambda_1}{\lambda_1+\lambda_2}\right)^{i} \left(\frac{\lambda_2}{\lambda_1+\lambda_2}\right)^{k+l-i} |k-i|.
\end{align}
From this equation, we can notice that the expected distance indeed exhibits a symmetry when the pairs $(\lambda_1, k)$ and $(\lambda_2, l)$ are interchanged.

We now substitute $p = \lambda_1/(\lambda_1+\lambda_2)$ to obtain the binomial form as follows:
\begin{eqnarray}
    \mathbb{E} [|x_k - y_l|] &=&  \frac{\lambda_1 + \lambda_2}{\lambda_1 \lambda_2}\sum_{i=0}^{k+l} \binom{k+l}{i} p^{i}(1-p)^{k+l-i} |k - i|\\
    &=&  \frac{\lambda_1 + \lambda_2}{\lambda_1 \lambda_2} \mathbb{E}_{i \sim P(i|k+l,p)}\left[|k - i|\right],
    \label{eqn:prop1_general}
\end{eqnarray} 
where the binomial distribution $P(i|k+l, p) = \binom{k+l}{i} p^{i}(1-p)^{k+l-i}$ is parameterized by $k+l$ and $p= \lambda_1/(\lambda_1+\lambda_2)$. The case of $l=k$ in \eqref{eqn:prop1_general} gives the desired \eqref{eqn:prop1}.

In minimizing the expected distance \eqref{eqn:prop1}, if $\frac{\lambda_1+\lambda_2}{2\lambda_1\lambda_2}$ is a constant, we can focus on minimizing the binomial expectation $f(p) = \mathbb{E}_{i \sim P(i|2k,p)}\left[|i - (2k - i)|\right]$. 
From the symmetry of $f(p) = f(1-p)$,  we obtain $\frac{\partial f}{\partial p}(p) = -\frac{\partial f}{\partial p}(1-p)$ hence $\frac{\partial f}{\partial p}(\frac{1}{2}) = 0$. 
From this, we establish that $p$ should be $\frac{1}{2}$, i.e., $\lambda_1$ must equal to $\lambda_2$, for the binomial expectation to reach its minimum.

The variance of $|x_k - y_k|$ can be obtained from its definition $\text{Var}[|x_k - y_k|] =\mathbb{E}[|x_k - y_k|^2] - \mathbb{E}[|x_k - y_k|]^2$. A straightforward algebra using the second moment of Erlang distributions yields 
\begin{align}
    \mathbb{E}[| x_k - y_l |^2] & = \mathbb{E}[ ( x_k - y_l )^2] = \mathbb{E}[ x_k^2] + \mathbb{E}[y_l^2] - 2 \mathbb{E}[x_k y_l] = \mathbb{E}[ x_k^2] + \mathbb{E}[y_l^2] - 2 \mathbb{E}[x_k] \mathbb{E}[y_l]\\
    & = \frac{k}{\lambda_1^2} + \left(\frac{k}{\lambda_1}\right)^2 + \frac{l}{\lambda_2^2} + \left(\frac{l}{\lambda_2}\right)^2 - 2 \frac{k}{\lambda_1} \cdot \frac{l}{\lambda_2} \\
    & = \frac{k}{\lambda_1^2} + \frac{l}{\lambda_2^2} + \left(\frac{k}{\lambda_1} - \frac{l}{\lambda_2}\right)^2.
\end{align}
Therefore the variance is obtained as
\begin{align}
    \text{Var}[| x_k - y_l |^2] = \frac{k}{\lambda_1^2} + \frac{l}{\lambda_2^2} + \left(\frac{k}{\lambda_1} - \frac{l}{\lambda_2}\right)^2 - \mathbb{E} [|x_k - y_l|]^2.
\end{align}
$\qedblack$   


\subsection{Proof of Proposition \ref{theorem:limiting}} \label{proof:limiting}

\textit{\textbf{Proof.}} We show $\lim_{k\rightarrow \infty} P(x_k < y_k) = 0$ so that the expectation $\lim_{k\rightarrow \infty} \mathbb{E}[|x_k - y_k|] = \mathbb{E}[x_k - y_k] = k(\frac{1}{\lambda_1} - \frac{1}{\lambda_2})$. The probability
\begin{eqnarray}
    \lim_{k\rightarrow \infty} P(x_k < y_k)
    &=& \lim_{k\rightarrow \infty} P\left( \frac{\lambda_2 y_k}{\lambda_2 y_k + \lambda_1 x_k} > \frac{\lambda_2}{\lambda_1+\lambda_2} \right) \\
    &=& \lim_{k\rightarrow \infty} P\left( z_k < \frac{\lambda_1}{\lambda_1+\lambda_2} \right),
\end{eqnarray}
where $z_k = \frac{\lambda_2 y_k}{\lambda_2 y_k + \lambda_1 x_k} \sim p(z_k|k,k)$ with the symmetric Beta density function $p(z_k|k,k)$. Here, by the Chebyshev's inequality, for any $\epsilon > 0$,
\begin{eqnarray}
    P\left(\Big|z_k-\frac{1}{2}\Big| > \epsilon\right) \le \frac{1}{\epsilon^2} \frac{1}{4(2k+1)}.
\end{eqnarray}
Thus,
\begin{eqnarray}
    \lim_{k\rightarrow \infty} P\left(\Big|z_k-\frac{1}{2}\Big| > \epsilon\right) = 0,
\end{eqnarray}
which shows the convergence in probability, i.e., $z_k \rightarrow 1/2$. Therefore, for $\lambda_1<\lambda_2$, $\lim_{k\rightarrow \infty} P\left( z_k < \frac{\lambda_1}{\lambda_1+\lambda_2} \right) = 0$, which shows 
\begin{align}
\lim_{k\rightarrow \infty}\mathbb{E}[|x_k - y_k|]  = \mathbb{E}[x_k - y_k] = k\left(\frac{1}{\lambda_1} - \frac{1}{\lambda_2}\right).    
\label{eqn:expectation_limit}
\end{align}
Dividing this equation by $k$ gives the expectation in \eqref{eqn:prop2}.

For the variance of $|x_k - y_k$, as $k\rightarrow\infty$,
\begin{align}
    \lim_{k\rightarrow \infty} \text{Var}[|x_k - y_k|] & = \lim_{k\rightarrow \infty} k\left(\frac{1}{\lambda_1^2} + \frac{1}{\lambda_2^2}\right) + \left(\frac{k}{\lambda_1} - \frac{k}{\lambda_2}\right)^2 - \mathbb{E} [|x_k - y_k|]^2 \label{eqn:var_limit1} \\
    & = k\left(\frac{1}{\lambda_1^2}+ \frac{1}{\lambda_2^2}\right),
    \label{eqn:var_limit2}
\end{align}
where the last two terms in \eqref{eqn:var_limit1} cancel out due to \eqref{eqn:expectation_limit}.
Dividing \eqref{eqn:var_limit2} by $k^2$ gives the variance in \eqref{eqn:prop2}.
$\hfill \qedblack$

\subsection{Derivation of the Expected Distance under a Support Shift} \label{appendix:temporal_shift}

Here we introduce a support shift $\Delta t \geq 0$ between two spikes $x_k, y_l$ from two different Poisson processes with rates $\lambda_1$, $\lambda_2$. 
The expectation $\mathbb{E}\left[|x_k+\Delta t - y_l|\right]$ is calculated in a similar manner to the previous derivations in Appendix~\ref{proof:expected_distance} as follows: 
\begin{equation}
    \mathbb{E}\left[|x_k+\Delta t - y_l|\right]=\int^\infty_{0}\int^\infty_{0}|x_k+\Delta t - y_l|\ p(x_k)p(y_l)dx_k dy_l, \label{eq:absolute_dif_general}
\end{equation}
where $p(x_k)$, $p(y_l)$ are Erlang distributions with rates $\lambda_1$, $\lambda_2$ and shapes $k, l$, respectively.

The double integration of \eqref{eq:absolute_dif_general} can be written as
\begin{eqnarray*}
    \mathbb{E}\left[|x_k+\Delta t - y_l|\right] 
    &=& \int^{\Delta t}_{0} \int^{\infty}_{0} (x_k + \Delta t - y_l) p(x_k)p(y_l)dx_k dy_l \\
    &+& \int^\infty_{\Delta t} \int^{y_l - \Delta t}_{0} -(x_k + \Delta t - y_l) p(x_k)p(y_l)dx_k dy_l \\
    &+& \int^\infty_{\Delta t} \int^{\infty}_{y_l - \Delta t} (x_k + \Delta t - y_l) p(x_k)p(y_l)dx_k dy_l.
\end{eqnarray*}
When calculating the above integral, the following equation is useful:
\begin{align}
    \int_a^b x^m (x+\Delta t)^n \exp(-\lambda x) dx 
    &= \int_a^b x^m \sum_{j=0}^n \left( \binom{n}{j} x^j \Delta t^{n-j} \right) \exp(-\lambda x) dx \nonumber\\
    &= \sum_{j=0}^n  \binom{n}{j} \Delta t^{n-j} \int_a^b x^{m+j} \exp(-\lambda x) dx  \nonumber\\
    &= \left. \sum_{j=0}^n \binom{n}{j} \Delta t^{n-j} \left( - \sum^{m+j}_{i=0}\frac{(m+j)!}{i!\lambda^{m+j-i+1}}x^i \exp(-\lambda x) \right)\right|_a^b. \label{eq:expon_int_2}
\end{align}
After a straightforward algebra using \eqref{eqn:erlang_x_k}, \eqref{eqn:erlang_y_l}, \eqref{eq:expon_int}, and \eqref{eq:expon_int_2}, we obtain 
\begin{align}
\begin{split}
    \mathbb{E}\left[|x_k+\Delta t - y_l|\right] 
    &= \left(\frac{k}{\lambda_1} - \frac{l}{\lambda_2} + \Delta t\right) 
    \left( 1 - 2\exp(-\lambda_2\Delta t) \sum_{j=0}^{l-1}\frac{(\lambda_2\Delta t)^j}{j!}\right) \\
    &+ 2\exp(-\lambda_2\Delta t) \left[ \sum_{i=0}^{k-1} \sum_{j=0}^{l-1} \binom{i+j}{i} \frac{(\lambda_2\Delta t)^{l-1-j}}{(l-1-j)!}  \cdot \frac{(k-i) \lambda_1^{i-1} \lambda_2^{j+1}}{(\lambda_1+\lambda_2)^{i+j+1}} \right. \left. +\Delta t \frac{(\lambda_2 \Delta t)^{l-1}}{(l-1)!}\right].
    \label{eqn:deriv_absolute_difference_general}
\end{split}
\end{align}
Note that, if $\Delta t < 0$, we can obtain $\mathbb{E}[|x_k - \Delta t - y_l|]$ by interchanging $(\lambda_1, k)$ and $(\lambda_2, l)$ in the above result since $\mathbb{E}[|x_k - \Delta t - y_l|] = \mathbb{E}[|y_l + \Delta t - x_k|]$. 

To gain a clearer understanding of the results, we explore several limiting cases. When $\Delta t \to \pm \infty$, the distance approaches to $\pm \left(\frac{k}{\lambda_1} - \frac{l}{\lambda_2} + \Delta t\right)$.
Conversely, when $\Delta t = 0$, we can observe that \eqref{eqn:deriv_absolute_difference_general} reduces to \eqref{eq:apen_intermediate}.

When $k=l=1$, we can organize the result as follows:
\begin{align}
    &\mathbb{E}\left[|x_1+\Delta t - y_1|\right] \nonumber \\
    &= \left(\frac{1}{\lambda_1} - \frac{1}{\lambda_2} + \Delta t\right) \left( 1 - 2\exp(-\lambda_2\Delta t) \right) + 2\exp(-\lambda_2\Delta t) \left( \frac{\lambda_2}{\lambda_1(\lambda_1+\lambda_2)} + \Delta t\right) \nonumber\\
    &= \exp(-\lambda_2 \Delta t)\left(\frac{\lambda_1 + \lambda_2}{\lambda_1 \lambda_2} \mathbb{E}_{i \sim P(i|2,p)}[|i-1|] \right) + \Delta t + (1-\exp(-\lambda_2\Delta t))\left(\frac{1}{\lambda_1} - \frac{1}{\lambda_2} \right),    \label{eq:deriv_absolute_difference_2}
\end{align}
which we have observed in \eqref{eq:deriv_absolute_difference}.

The variance of $|x_k + \Delta t - y_k|$ can be obtained from its definition $\text{Var}[|x_k + \Delta t - y_k|] =\mathbb{E}[|x_k + \Delta t - y_k|^2] - \mathbb{E}[|x_k + \Delta t - y_k|]^2$. A straightforward algebra yields 
\begin{align*}
    \mathbb{E}[| x_k + \Delta t - y_l |^2] & = \mathbb{E}[ ( x_k + \Delta t - y_l )^2] \\
    & = \mathbb{E}[ x_k^2] + \mathbb{E}[y_l^2] - 2 \mathbb{E}[x_k y_l] + \Delta t^2 + 2\Delta t (\mathbb{E}[x_k] - \mathbb{E}[y_l])\\
    & = \frac{k}{\lambda_1^2} + \frac{l}{\lambda_2^2} + \left(\frac{k}{\lambda_1} - \frac{l}{\lambda_2}\right)^2 + \Delta t^2 +2\Delta t \left(\frac{k}{\lambda_1} - \frac{l}{\lambda_2}\right).
\end{align*}
Therefore the variance is obtained as
\begin{align*}
    \text{Var}[| x_k + \Delta t - y_l |^2] = \frac{k}{\lambda_1^2} + \frac{l}{\lambda_2^2} + \left(\frac{k}{\lambda_1} - \frac{l}{\lambda_2}\right)^2  + \Delta t^2 + 2\Delta t \left(\frac{k}{\lambda_1} - \frac{l}{\lambda_2}\right) - \mathbb{E} [|x_k + \Delta t - y_l|]^2.
\end{align*}

\section{Numerical Validation of Derived Results}
\label{appendix:numerical}

\begin{figure}[t]
\centering
\begin{subfigure}{.47\textwidth}
    \centering
    \includegraphics[width=1.0\textwidth]{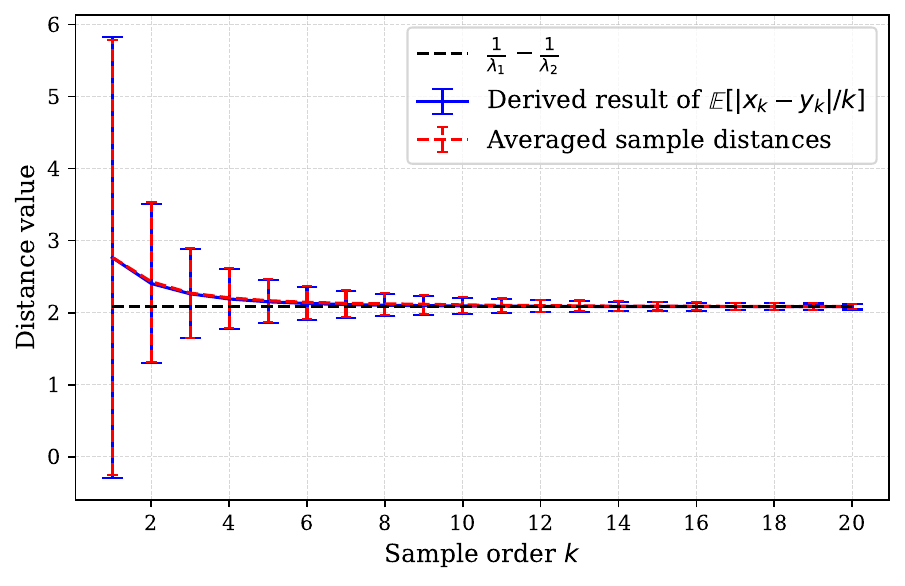}
    \caption{The mean and standard deviation of $\frac{|x_k - y_k|}{k}$}
\end{subfigure}~
\begin{subfigure}{.52\textwidth}
    \centering
    \includegraphics[width=1.0\textwidth]{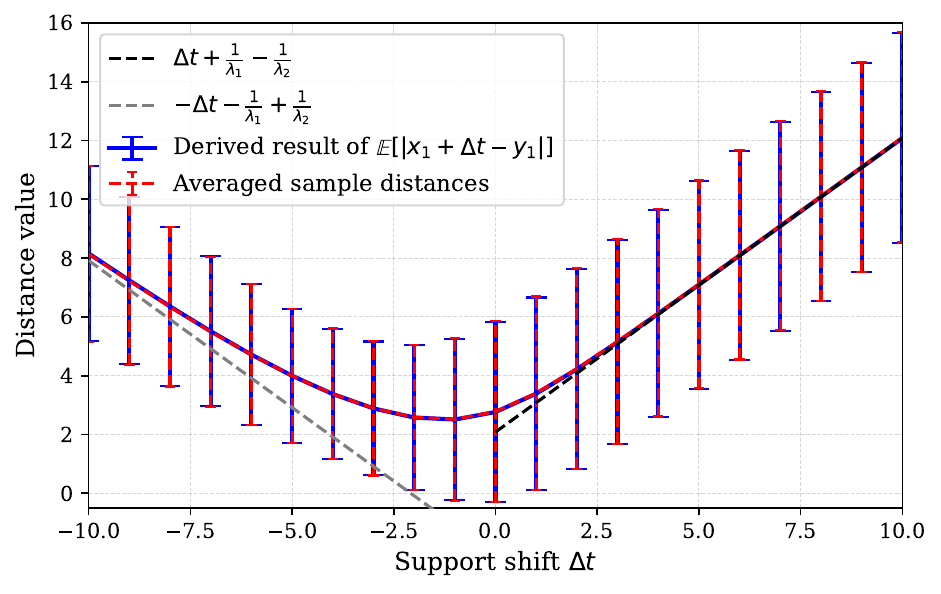}
    \caption{The mean and standard deviation of $\mathbb{E}[|x_1 + \Delta t - y_1|]$}
\end{subfigure}
\caption{Numerical analyses on  $\mathbb{E}[|x_k-y_k|]$ and $\mathbb{E}[|x_1 + \Delta t - y_1|]$. 
We have used $\lambda_1 = 0.3, \lambda_2 = 0.8$ and repeated $20,000$ trials to obtain the mean and standard deviation of sample distances. 
In (a), $\mathbb{E}[|x_k-y_k|/k]$ converges to the rate difference $(1/\lambda_1 - 1/\lambda_2)$ as the event order $k$ increases. 
In (b), as $\Delta t \to \pm \infty$, the expected distance converges to $\pm(\Delta t + 1/\lambda_1 - 1/\lambda_2)$.
}
\label{fig:properties_of_deriv}
\end{figure}

We conduct numerical experiments to verify the derivations in Section~\ref{section:main} and Appendix~\ref{appendix:derivations}. 
Figure~\ref{fig:properties_of_deriv}(a) displays the mean and standard deviation of the distances between spikes (normalized by the sample order $k$) from Poisson processes with fixed rates $\lambda_1$ and $\lambda_2$, as a function of $k$. 
As can be seen from the figure, our theoretical predictions in \eqref{eqn:prop1} and the variance of $|x_k - y_k$, depicted as blue solid lines, match well with the results from samples depicted in red dashed lines.
Furthermore, it is evident that as $k$ increases, the normalized distance $|x_k - y_k|/k$ converges towards $1/\lambda_1 - 1/\lambda_2$. 
In Figure~\ref{fig:properties_of_deriv}(b), blue solid lines correspond to our theoretical derivations for the mean and standard deviation of $|x_1 + \Delta t - y_1|$ in \eqref{eq:deriv_absolute_difference}. 
The results match those from samples depicted in red dashed lines.

\section{Extending the Derived Results to Sliced Wasserstein Distances}
\label{appendix:sliced_wasserstein}

Our analytic derivations—particularly those showing how the one-dimensional Wasserstein distance captures pointwise density and support differences—can be extended beyond the Poisson setting. 
Specifically, the framework developed in Section~\ref{sec:time_varying} for time-varying rate Poisson processes provides a natural bridge to understanding sliced Wasserstein distances, which are widely used in practical applications.

The sliced Wasserstein distance is defined as the expectation of the one-dimensional Wasserstein distance applied to projections of multidimensional distributions along arbitrary directions \cite{bonneel2015sliced}.
When high-dimensional samples are projected onto random one-dimensional (1D) subspaces, the projected samples can be interpreted as arrival times from Poisson processes whose time-varying rates are proportional to the projected 1D probability density functions (PDFs). 
This interpretation allows us to apply our analytic framework to characterize how the one-dimensional Wasserstein distances along each projection direction capture pointwise density and support differences.

Given that sliced Wasserstein distance aggregates these directional differences across multiple random projections, our derivations can offer a theoretical pathway for understanding its information-processing characteristics.
To demonstrate the applicability and tractability of this extension, we now consider an illustrative example.

\paragraph{An illustrative example.}
Suppose two distinct two-dimensional (2D) distributions with piecewise constant PDFs, defined over regions separated by piecewise linear boundaries. 
When both are projected onto the same one-dimensional (1D) subspace, the resulting marginal PDFs become piecewise linear.
This structure generalizes naturally to higher dimensions.

We model the projected distributions using time-varying rate Poisson processes with (piecewise linear) rate functions $\mu(t)$ and $\nu(t)$, proportional to the projected PDFs of the two distributions, respectively.
Their cumulative intensity functions are given by
\[
m(x) = \int_0^x \mu(t)\,dt, \quad 
n(x) = \int_0^x \nu(t)\,dt,
\]
which are piecewise quadratic. 
The corresponding inverse cumulative intensity functions take the forms
\[
m^{-1}(u) = \sqrt{a_i u + b_i} + c_i, \quad \text{for} \quad u \in [U_i, U_{i+1}),
\quad
n^{-1}(v) = \sqrt{\tilde{a}_j v + \tilde{b}_j} + \tilde{c}_j, \quad \text{for} \quad v \in [V_j, V_{j+1}),
\]
where the constants $a_i, b_i, c_i$, $\tilde{a}_j, \tilde{b}_j, \tilde{c}_j$, and the intervals $[U_i, U_{i+1})$ (for $i = 1, \ldots, I$) and $[V_j, V_{j+1})$ (for $j = 1, \ldots, J$) are determined by the original 2D distributions and the projection direction.

Under this construction, the expected squared distance between projected samples $\mathbb{E}[(x_k - y_l)^2]$ (which corresponds to replacing the absolute difference in \eqref{eqn:time_varying} with its square) decomposes into a sum of integrals over all pairs of intervals:
\begin{align}
\sum_{i=1}^I \sum_{j=1}^J \int_{V_j}^{V_{j+1}} \int_{U_i}^{U_{i+1}} \left( 
\sqrt{a_i u + b_i} + c_i - 
\sqrt{\tilde{a}_j v + \tilde{b}_j} - \tilde{c}_j 
\right)^2 e^{-u} u^{k-1} e^{-v} v^{l-1}\, du\, dv.
\label{eqn:sliced_wass_example}
\end{align}
Each integral is analytically tractable using identities involving the incomplete Gamma function. 
Even when the Gamma function arguments fall outside standard domains (e.g., negative values), the integrals remain numerically stable and can be efficiently evaluated using algorithms such as Thompson’s method~\cite{thompson2013algorithm}.

A complete analytic treatment of the sliced Wasserstein distance can be obtained by aggregating expressions of the form in \eqref{eqn:sliced_wass_example} across multiple random projection directions.
This example illustrates that extending our framework to the sliced Wasserstein setting is not only conceptually natural but also analytically and computationally tractable.

\section{Details of the Synthetic Data Experiment}
\label{appendix:synthetic}
For the synthetic example in Figure~\ref{fig:simple_example} of Section~\ref{sec:synthetic}, we set $\mu(t) = \lambda$ for $t\in [0,1]$ and zero otherwise. 
The rate $\nu(t)$ is set using $(r, \Delta t)\in \RE^2$ as $\nu(t) = r\lambda$ for $t\in [\Delta t, \Delta t + 1/(r+1))$, $\lambda/r$ for $t \in [\Delta t + 1/(r+1), \Delta t + 1]$, and zero otherwise, where $r \in [e^{-2}, e^2]$ is the rate scaling variable and $\Delta t \in [-2, 2]$ is the support shift variable.
This configuration ensures that the expected number of samples is consistent across both processes. 
The baseline rate is set to $\lambda = 100$, resulting in an expected count of 100 samples for the given rates. 
In the experiments, the JS divergence is computed between probability mass functions based on ten equal-sized bins.

For the rate and support difference prediction task in Section~\ref{sec:synthetic}, we set $\mu(t) = \lambda$ for $t\in [0,1]$ and zero otherwise. 
The rate $\nu(t)$ is defined using $(r_1, r_2, \Delta t)\in \RE^3$ as $\nu(t) = r_1 \lambda$ for $t\in [\Delta t, \Delta t + 1/(2r_1))$, $r_2 \lambda$ for $t \in [\Delta t + 1/(2r_1), \Delta t + 1/(2r_1) + 1/(2r_2)]$, and zero otherwise, where $r_1, r_2 \in [e^{-2}, e^2]$ are the rate scaling variables and $\Delta t \in [-2, 2]$ is the support shift variable.
The baseline rate is set to $\lambda = 100$.\\
To construct the synthetic training set, for the $i$-th training data, we sample $\log r_{1,i}$, $\log r_{2,i}$, and $\Delta t_i$ uniformly from $\mathcal{U}[-2, 2]$ to define $\nu(t)$.
Using these, we generate Poisson process sequences $X_{i} = \{x_{i,k}\}_{k=1}^{T_{\mu,i}}$ and $Y_{i} = \{y_{i,k}\}_{k=1}^{T_{\nu,i}}$ according to $\mu(t)$ and $\nu(t)$, respectively. 
Repeating this process $N = 10,000$ times yields the training dataset $\{(X_{i}, Y_{i}, z_i)\}_{i=1}^N$, where $z_i = (\log r_{1, i}, \log r_{2, i}, |\Delta t_i|) \in \RE^3$ serves as the ground truth for prediction. 
A similarly generated dataset is used for testing.\\
The sample transport cost feature for two given sequences $X = \{x_{k}\}_{k=1}^{T_\mu}$ and $Y = \{y_{k}\}_{k=1}^{T_\nu}$ is defined as $c = (c_1, \ldots, c_{10}) \in \mathbb{R}^{10}$, where each decile's transport cost is given by $c_i = \int_{0.1 \cdot (i-1)}^{0.1 \cdot i} \left| \hat{P}^{-1}(u) - \hat{Q}^{-1}(u) \right| du$. 
Here, $\hat{P}$ and $\hat{Q}$ represent the cumulative distribution functions (CDFs) of the empirical measures derived from the sequences $X$ and $Y$, respectively.
The bin-wise JS-divergence features between $X$ and $Y$ are defined as $(v_1, \ldots, v_{10}) \in \mathbb{R}^{10}$, where $v_k = \frac{1}{2} \left( P_k \log \left(\frac{P_k}{M_k}\right) + Q_k \log \left(\frac{Q_k}{M_k}\right) \right)$, with $P_k$ and $Q_k$ denoting the probability masses of $X$ and $Y$ in the $k$-th bin of ten equal-sized bins, respectively, and $M_k = \frac{P_k + Q_k}{2}$.
The directed Hausdorff distance features between $X$ and $Y$ are defined as $(\sup_{k=1}^{T_{\mu}} \inf_{l=1}^{T_{\nu}} |x_{k} - y_{l}|, \sup_{l=1}^{T_{\nu}} \inf_{k=1}^{T_{\mu}} |x_{k} - y_{l}|) \in \mathbb{R}^2$.\\
To estimate $\log r_1, \log r_2$, and $|\Delta t|$ from the features, we use a three-layer fully connected neural network with 64 hidden nodes per layer and ReLU activation functions. 
The model is trained to minimize the mean squared error loss on 10,000 training data, using the Adam optimizer with a learning rate of 0.001 and batch size of 10,000 for 10,000 epochs. 
The mean and standard error of $R^2$ on 10,000 test data over five random seeds are reported in Table~\ref{table:synthetic}.

\section{Details of Retinal Ganglion Cell Stimulus Type Classification}
\label{appendix:spikebench}

\begin{figure}[ht!]
\centering
\includegraphics[width=.9\textwidth]
{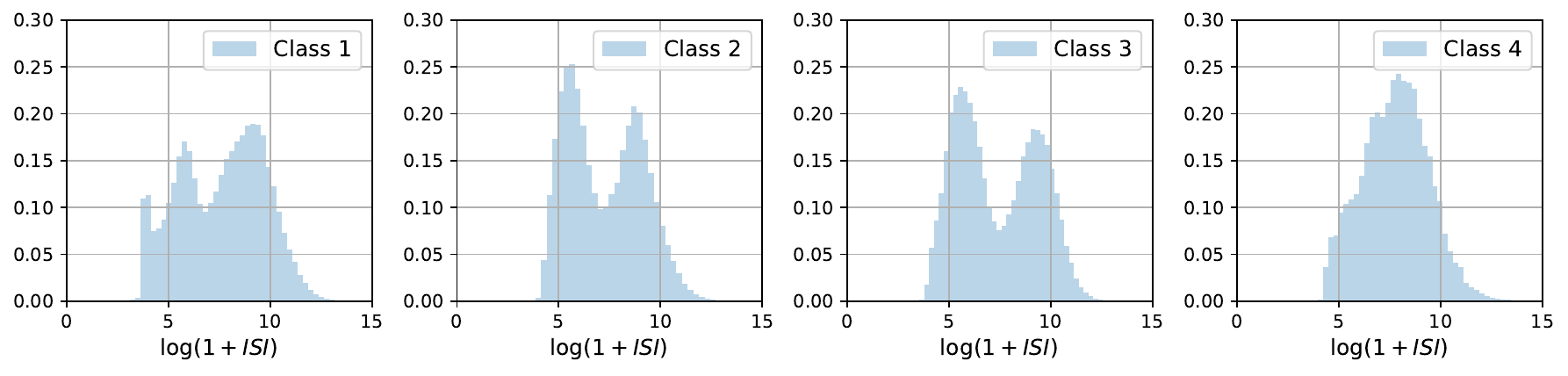}
\caption{The normalized histograms of $\log(1+ISI)$ values for each class in the retinal ganglion cell spike train data.}
\label{fig:isi_distribution}
\end{figure}

We use spike time data from multi-electrode array recordings of salamander retinal ganglion cells under four stimulus conditions, as considered in \cite{prentice2016error}: a white noise checkerboard (class 1), a repeated natural movie (class 2), a non-repeated natural movie (class 3), and a bar exhibiting random one-dimensional motion (class 4). 
Following the data preprocessing method in \cite{lazarevich2023spikebench}, we extract ISI values from single neurons using a sliding window of size 200 with a step of 100 ISIs, producing 200-dimensional temporal vectors. 
Applying this process across all neurons results in the dataset $\mathcal{D}=\{(X_i, y_i)\}_{i=1}^T$, where $X_i = (x_{i,1}, \ldots, x_{i,200}) \in \RE^{200}$, $y_i \in \{1,2,3,4\}$, $i=1,\ldots,T$, and $T$ is the total number of data. \\
The dataset is split into training and test sets, ensuring no neuron appears in both.
For the multiclass classification problem, the training set contains 6,280, 7,619, 2,753, and 1,705 samples for classes 1, 2, 3, and 4, respectively, while the test set contains 2,730, 3,260, 1,139, and 941 samples. 
In total, 18,357 training and 8,070 test samples are used. 
The ISI value distribution in the training data is shown in Figure~\ref{fig:isi_distribution}.

The sample transport cost feature in the SD1 approach is computed by comparing the empirical measure of each data point $X_i=\{x_{i,k}\}_{k=1}^{200}$ with that of all training samples $Y=\{x_{i,k}\}_{i=1}^{T_{train}},_{k=1}^{200}$.
Letting $D$ be the feature dimension, the sample transport distance feature is defined as $c=(c_1, \ldots, c_{D}) \in \RE^{D}$, where the $i$-th feature is defined as $c_i = \int_{(i-1)/D}^{i/D} \left| \hat{P}^{-1}(u) - \hat{Q}^{-1}(u) \right| du$, with $\hat{P}$ and $\hat{Q}$ as the CDFs of the empirical measures for $X$ and $Y$, respectively.
In the SD2 approach, features for each class are computed similarly by redefining $Y$ and $\hat{Q}$ for each class, and the transport costs from all classes are combined into a feature set. 
In the experiments, we use $D = 200$.

We use four 1D CNN models: FCN and ResNet for time series \cite{wang2017time}, InceptionTime \cite{ismail2020inceptiontime}, and XceptionTime \cite{rahimian2019xceptiontime}, implemented in the \texttt{tsai} library \cite{tsai}.\\
When incorporating the sample transport distance feature, we employ separate CNNs of the same architecture to process the ISI data and the sample transport distance feature. 
This approach is adopted instead of directly concatenating these features along the channel dimension. 
Our design choice stems from the fundamental difference in the semantic meaning of their indices: while each index in the ISI vector corresponds to a time step, the indices in SD1 and SD2 represent specific quantile intervals. 
Concatenating them as a single input channel would therefore be semantically inconsistent and {\it ad hoc}. 
Consequently, we process each input stream independently and concatenate their respective outputs prior to the final classification layer.

Both ISI values and features are transformed with the $\log(1+x)$ function and standardized before being fed into the models.
For SD2, sample transport distance features from each class aggregation are combined into a multi-channel input. 
The CNN output dimensions are set to 128 for ISI data and 64 for the sample transport distance feature, and these outputs are concatenated for final classification.\\
Models are trained using \texttt{pytorch} \cite{paszke2019pytorch} with SGD for 200 epochs, a batch size of 256, and an initial learning rate of 0.1, which decays to 0 in the last 1/4 of training using cosine annealing.

\begin{table}[t]
\caption{Classification test AUC for retinal stimulus types (expanded).}
\label{table:retina_expanded}
\begin{center}
\resizebox{\columnwidth}{!}{%
\begin{sc}
\begin{tabular}{lccccccc}
\toprule
Method & Retina-All & Retina14 & Retina23 & Retina12 & Retina34 & Retina13 & Retina24 \\
\midrule
FCN & 0.945 $\pm$ 7e-04 & 0.962 $\pm$ 7e-04 & 0.925 $\pm$ 3e-04 & 0.955 $\pm$ 7e-04 & 0.978 $\pm$ 6e-04 & 0.910 $\pm$ 6e-04 & 0.988 $\pm$ 1e-04 \\
FCN + SD1 & {\bf 0.951} $\pm$ 4e-04 & {\bf 0.971} $\pm$ 9e-04 & {\bf 0.931} $\pm$ 6e-04 & 0.957 $\pm$ 2e-03 & {\bf 0.983} $\pm$ 4e-04 & {\bf 0.934} $\pm$ 5e-04 & 0.988 $\pm$ 9e-05 \\
FCN + SD2 & 0.945 $\pm$ 1e-03 & {\bf 0.968} $\pm$ 1e-03 & {\bf 0.935} $\pm$ 4e-04 & {\bf 0.958} $\pm$ 1e-03 & 0.978 $\pm$ 5e-04 & {\bf 0.927} $\pm$ 8e-04 & 0.987 $\pm$ 4e-04 \\
\midrule
InceptonTime & 0.937 $\pm$ 8e-04 & 0.955 $\pm$ 2e-03 & 0.889 $\pm$ 2e-03 & 0.932 $\pm$ 2e-03 & 0.982 $\pm$ 1e-03 & 0.897 $\pm$ 9e-04 & 0.985 $\pm$ 9e-04 \\
InceptonTime + SD1 & {\bf 0.951} $\pm$ 6e-04 & 0.960 $\pm$ 2e-03 & {\bf 0.912} $\pm$ 3e-03 & {\bf 0.955} $\pm$ 2e-03 & 0.981 $\pm$ 9e-04 & {\bf 0.914} $\pm$ 2e-03 & {\bf 0.988} $\pm$ 7e-04 \\
InceptonTime + SD2 & {\bf 0.950} $\pm$ 1e-03 & {\bf 0.966} $\pm$ 2e-03 & {\bf 0.913} $\pm$ 1e-03 & {\bf 0.954} $\pm$ 2e-03 & 0.980 $\pm$ 8e-04 & {\bf 0.906} $\pm$ 4e-03 & {\bf 0.990} $\pm$ 4e-04 \\
\midrule
ResNet & 0.937 $\pm$ 7e-04 & 0.967 $\pm$ 2e-03 & 0.898 $\pm$ 8e-04 & 0.941 $\pm$ 1e-03 & 0.984 $\pm$ 1e-03 & 0.892 $\pm$ 3e-03 & 0.984 $\pm$ 7e-04 \\
ResNet + SD1 & {\bf 0.943} $\pm$ 1e-03 & 0.965 $\pm$ 8e-04 & {\bf 0.912} $\pm$ 1e-03 & {\bf 0.948} $\pm$ 2e-03 & {\bf 0.988} $\pm$ 4e-04 & {\bf 0.907} $\pm$ 3e-03 & {\bf 0.990} $\pm$ 5e-04 \\
ResNet + SD2 & {\bf 0.948} $\pm$ 5e-04 & 0.967 $\pm$ 2e-03 & {\bf 0.913} $\pm$ 2e-03 & {\bf 0.949} $\pm$ 2e-03 & 0.982 $\pm$ 7e-04 & {\bf 0.912} $\pm$ 2e-03 & {\bf 0.989} $\pm$ 1e-03 \\
\midrule
XceptionTime & 0.944 $\pm$ 6e-04 & 0.970 $\pm$ 1e-03 & 0.930 $\pm$ 8e-04 & 0.961 $\pm$ 9e-04 & 0.984 $\pm$ 6e-04 & 0.911 $\pm$ 1e-03 & 0.982 $\pm$ 8e-04 \\
XceptionTime + SD1 & 0.947 $\pm$ 2e-03 & {\bf 0.979} $\pm$ 6e-04 & 0.932 $\pm$ 1e-03 & {\bf 0.963} $\pm$ 5e-04 & 0.982 $\pm$ 8e-04 & {\bf 0.916} $\pm$ 2e-03 & 0.982 $\pm$ 9e-04 \\
XceptionTime + SD2 & {\bf 0.950} $\pm$ 6e-04 & {\bf 0.978} $\pm$ 9e-04 & 0.932 $\pm$ 1e-03 & 0.960 $\pm$ 1e-03 & 0.977 $\pm$ 1e-03 & 0.912 $\pm$ 1e-03 & 0.977 $\pm$ 2e-03 \\
\bottomrule
\end{tabular}
\end{sc}
}
\end{center}
\end{table}

In Tables~\ref{table:retina} and \ref{table:retina_expanded}, we report the mean and standard error of test AUC over five random seeds for each classification task. 
In the tables, ``Retina-All" denotes the multiclass classification setting distinguishing all classes, while in binary classification, the two numbers following ``Retina" indicate the included classes (e.g., Retina14 distinguishes class 1 from class 4).
For the multiclass problem, the one-vs-rest AUC values obtained for each class are weighted-averaged. 
A one-sided t-test was performed, with p-values below 0.05 highlighted in bold.
Table~\ref{table:retina_expanded} extends the main text results by also including binary classification outcomes for Retina12, Retina34, Retina13, and Retina24. 
The results show that incorporating the sample transport distance feature further improves stimulus type classification performance in most cases.

\section{Details of Human Neural Spike Train Analysis}
\subsection{Description of Human Neural Spike Train Data}
\label{appendix:neural_spike}
\begin{figure}[ht!]
\centering
\includegraphics[width=.7\textwidth]
{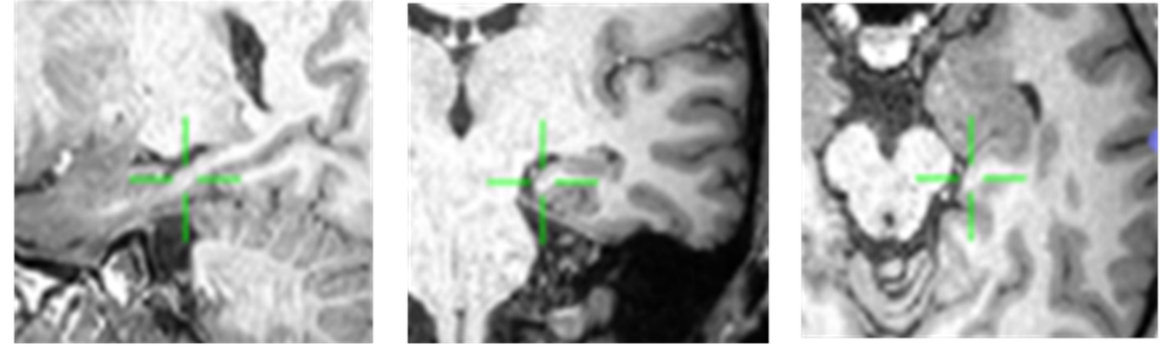}
\caption{Hippocampal microelectrodes are localized using structural MRI images, with green arrows indicating insertion into the right hippocampus in sagittal, coronal, and axial views from left to right.}
\label{fig:hippocampal_MRI}
\end{figure}

\begin{figure}[ht!]
\centering
\includegraphics[width=.9\textwidth]
{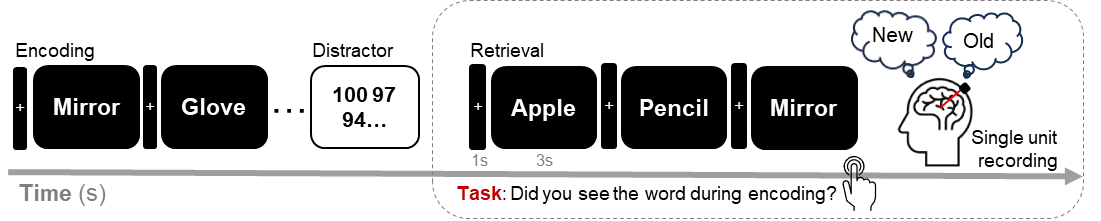}
\caption{Example of the timeline of the visual item memory task.}
\label{fig:retrieval_task}
\end{figure}

We have used the neural activity data obtained from a human subject performing a word memory task in \cite{jun2020task, jun2023hippocampal}. 
Extracellular neural activity was recorded using 40-\textmu m micro-wires implanted in a depth electrode and inserted in the hippocampus as shown in Figure~\ref{fig:hippocampal_MRI}. 
Spiking data were recorded using a 32-channel Neuralynx ATLAS system and acquired by bandpass filtering the raw signal from 600 to 9000Hz with a sampling rate of 32kHz. Spike sorting and preprocessing were performed using Wave clus \cite{doi:10.1152/jn.00339.2018}.

The word memory task comprised three successive stages of encoding, a distractor period, and retrieval, as shown in Figure~\ref{fig:retrieval_task}. 
During encoding, the subject performed 120-word items of one concrete noun, randomly shown one at a time.
A word was visible for 4s, followed by a white fixation with a black screen of 1s. 
Following the final word of the encoding block, the subject took a 10-minute break and performed a 30-second math distractor task.

In our experiments, we have used the neural activity obtained in the retrieval stage, where the subject was to respond whether the word had been presented before or was new or the subject was not sure.
The words were presented to the subject every 3.5 seconds, and the timing of the words presented to the subject during the task was also recorded, along with the subject's responses.
We segment the entire recording of 667-second into a series of 140-second windows with a 3.5-second sliding interval between each window.
Before segmenting the entire recording, we processed this data using spike sorting techniques \cite{doi:10.1152/jn.00339.2018}. 

\subsection{Isomap Embeddings of Human Neural Spike Trains with Different Distance Metrics}
\label{appendix:embedding}
\begin{figure*}[htb]
\centering
\begin{subfigure}{.31\textwidth}
\centering
\includegraphics[width=1.0\textwidth]{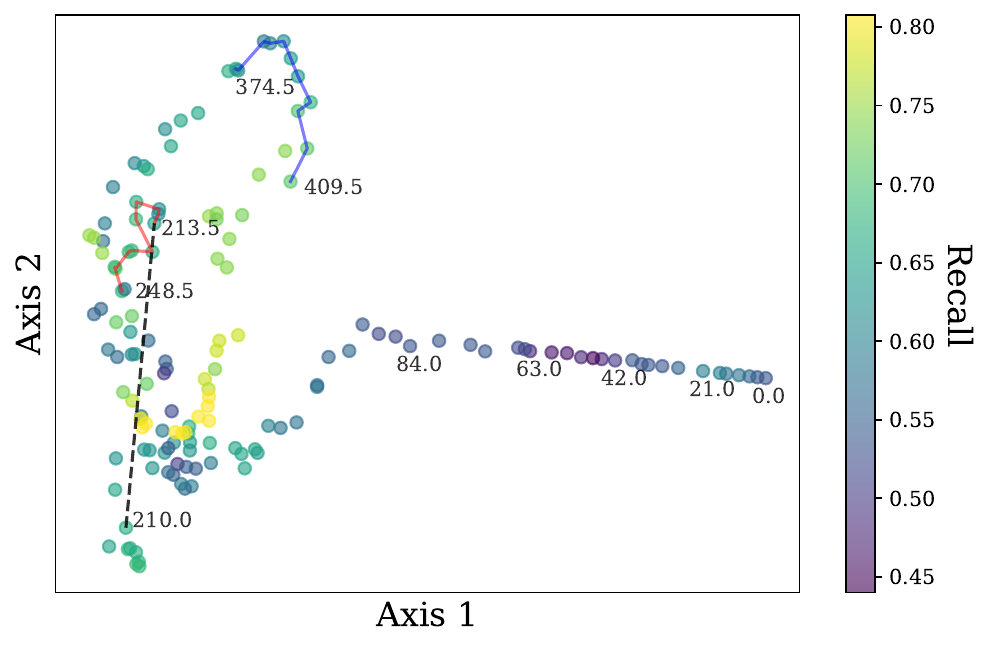}
\caption{VP distance ($q=1e-5$)}
\end{subfigure} \quad
\begin{subfigure}{.31\textwidth}
\centering
\includegraphics[width=1.0\textwidth]{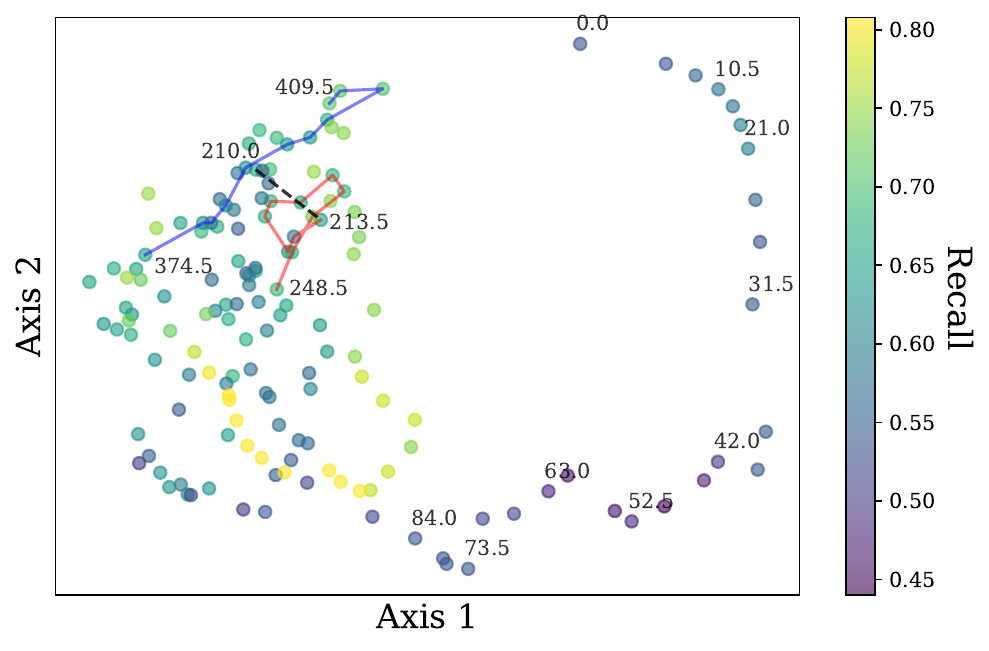}
\caption{KFS distance ($\tau=1e4$)}
\end{subfigure} \quad
\begin{subfigure}{.31\textwidth}
\centering
\includegraphics[width=1.0\textwidth]{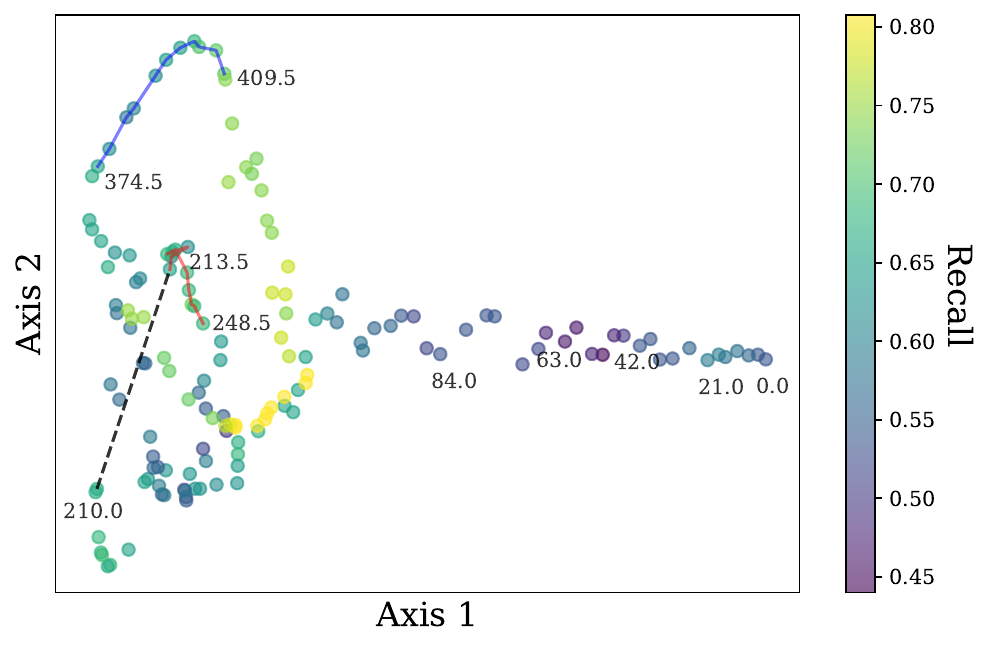}
\caption{KFS distance ($\tau=1e5$)}
\end{subfigure}
\caption{Three Isomap embeddings of human neural spike trains. We present the embedding obtained using the Victor-Purpura (VP) distance with a spike time-shift cost $q=1e-5$ in (a), that from a kernel feature space (KFS) distance with bandwidth parameters $\tau=1e4$ and $\tau=1e5$ in (b) and (c), respectively. 
}
\label{fig:neural_embedding_appendix}
\end{figure*}
The Victor-Purpura (VP) distance \cite{victor1997metric} defines the distance between spike trains as the minimum cost required to modify one spike train into another with a spike time-shift cost of $q$. 
When $q$ is 0, the distance becomes the difference in the number of spikes. 
As $q$ increases, the distance reflects the temporal difference information between spikes.
We consider the composite distance measure of the dimension-wise VP distances with $q=1e-4$ computed alike the composite Wasserstein distance in Section~\ref{sec:spike_train}.
In Figure~\ref{fig:neural_embedding_appendix}(a), we display the embedding obtained using the VP distance with $q=1e-5$. (We note that further increasing $q$ did not lead to meaningful embeddings.)

We additionally consider a kernel feature space (KFS) distance using a kernel defined in \cite{park2013kernel}. 
The distance measure in the feature space is calculated as $\sqrt{k(x,x) - 2k(x,y) + k(y,y)}$ for a kernel $k(x,y)$, where the kernel between spike trains $\{x_i\}_{i=1}^N$ and $\{y_j\}_{j=1}^M$ is defined as $k(\{x_i\}_{i=1}^N,\{y_j\}_{j=1}^M) = \sum_{i=1}^N \sum_{j=1}^M \exp\left(-\frac{1}{\tau}|x_i - y_j|\right)$ with $\tau > 0$ as the bandwidth parameter of the kernel.
If $\tau$ is very large (i.e., $\tau \rightarrow \infty$, the feature space distance between two spike trains becomes their spike count difference, and when the bandwidth is moderate, it can reflect temporal differences.
We consider the composite distance measure of the dimension-wise KFS distances calculated similarly to the composite Wasserstein distance in Section~\ref{sec:spike_train}.
We show the embeddings obtained for the kernel with bandwidth parameters $\tau=1e4$ and $\tau=1e5$ in Figures~\ref{fig:neural_embedding_appendix}(b) and (c), respectively.
(A further decrease in $\tau$ did not lead to meaningful embeddings.)

When closely observing the embeddings, the embeddings in Figures~\ref{fig:neural_embedding_appendix}(a) and (c) are qualitatively similar to that of the spike count difference in Figure~\ref{fig:neural_embedding}(b), showing the primary variation confined to distinguish the initial windows from the others.
Furthermore, the embedding in Figure~\ref{fig:neural_embedding_appendix}(b) is qualitatively similar to that in Figure~\ref{fig:neural_embedding}(c), and two consecutive windows with a large rate difference (corresponding to the first two spike histograms in Figure~\ref{fig:neural_spike_data}), connected by a black dotted line, are quite close to each other, indicating that this information is not effectively captured in these cases.
Therefore, in settings like ours where the goal is to obtain an embedding that accurately reflects the temporal patterns of spike trains, the Wasserstein distance can be more effective than the VP and KFS distances in capturing the rate differences and temporal shifts inherent in spike trains.

\section{Details of Amino Acid Contact Analysis}
\label{appendix:amino_acid}
\begin{figure*}[ht]
\centering
\includegraphics[width=0.78\textwidth]{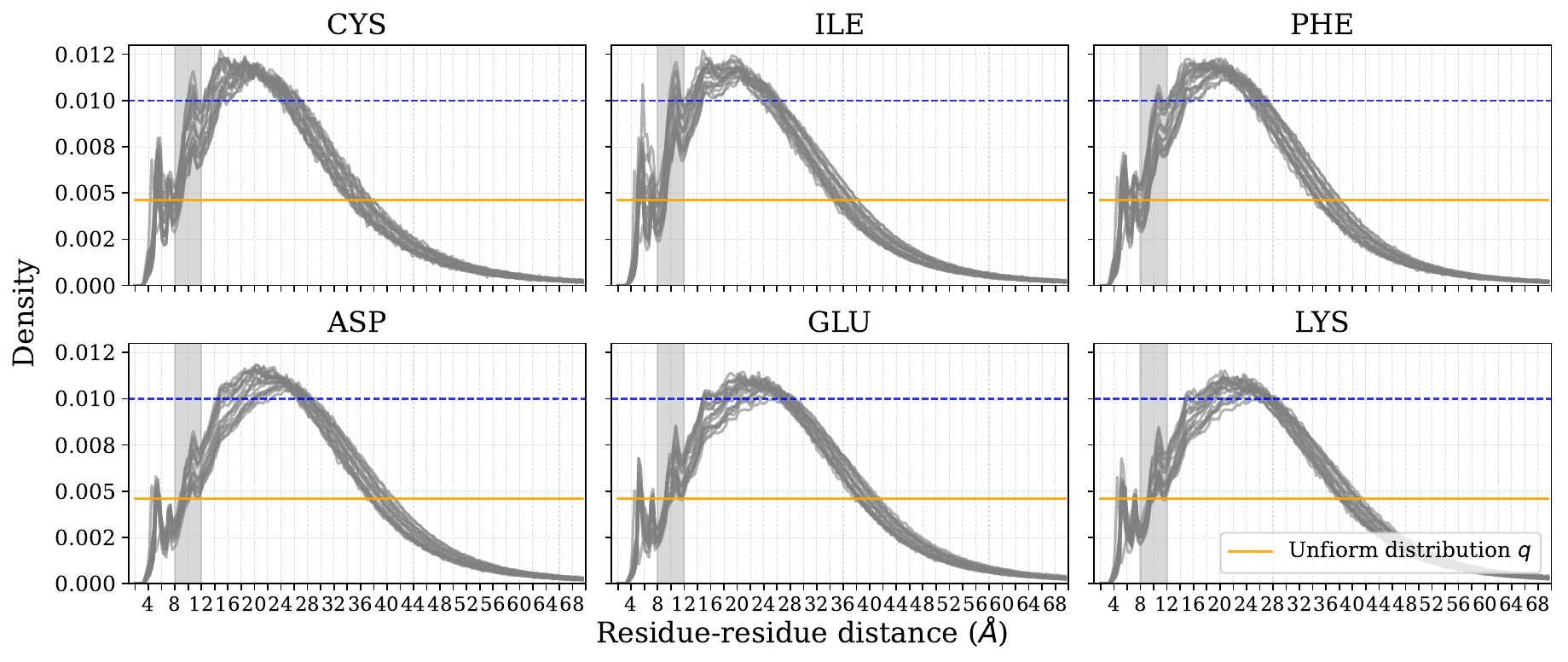}
\caption{The region of significant rate difference in distributions $p_{ij}$. 
The multiple gray curves depict each $p_{ij}$ for all the other $j\neq i$ for the given amino acid $i$.
The amino acids in the first row (CYS, ILE, and PHE) exhibit notably larger density differences from $p_{ij}$ to $q$ in the range of 8-12\AA\ compared to those in the second row.}
\label{fig:protein_distributions}
\end{figure*}
To represent amino acid contact frequencies, we measured pairwise straight-line distances between amino acids in protein structures from the PDB database\footnote{The PDB database (\texttt{https://www.rcsb.org/}), accessed as of February 2022, was used.}. 
For each protein in our dataset, we collected distances between all amino acid pairs within the same protein, yielding 190 histograms corresponding to the $\binom{20}{2}$ possible amino acid combinations. 
Due to the large number of pairs ($\sim$0.32 billion), we applied quantization to simplify the computation.
 
The contact frequencies in naturally occurring proteins (e.g., the PDB dataset we used) generally exhibit similar patterns, as shown in Figure~\ref{fig:protein_distributions}. 
To analyze these patterns, we compare the contact frequencies $p_{ij}$ to a uniform distribution $q$, a simple yet effective approach. 
Since it is challenging to directly interpret $W(p_{ij}, q)$ or $D_{KL}(p_{ij}||q)$ as the dissimilarity between the contact frequencies of amino acids $i$ and $j$, we scaled the values to range between 0.4 and 1.4, which enabled the meaningful embeddings presented in Figure~\ref{fig:protein_map}.
We found that varying the scaling within a range that maintained the ratio between minimum and maximum values between 3 and 10 did not substantially alter the qualitative results.
This approach proved effective in capturing the subtle differences observed in amino acid contact frequencies.

To obtain the embeddings, we employed the Riemannian geometric manifold learning algorithm from \cite{jang2021riemannian}, which is designed to effectively preserve the local geometry of the data. 
This method involves solving a non-convex optimization problem for the embeddings. 
We used the Isomap embeddings (using five nearest neighbors to construct geodesic distances) as the initial guess and selected other hyperparameters as described in \cite{jang2021riemannian}.
\end{document}